\documentclass[review]{elsarticle}
\graphicspath{ {./figures/} }
\usepackage{hyperref}
\usepackage{float}
\usepackage{verbatim} 
\usepackage{apalike}
\restylefloat{figure}
\floatstyle{plaintop} 
\restylefloat{table}

\usepackage[T1]{fontenc}
\usepackage{graphicx}
\usepackage{booktabs}
\usepackage{amsmath}
\usepackage[misc]{ifsym}

\usepackage[T1]{fontenc}
\usepackage{booktabs}
\usepackage[misc]{ifsym}

\usepackage{dsfont}
\usepackage{amsmath,amssymb,amsfonts}
\usepackage{hyperref}

\usepackage{algorithm}
\usepackage{algorithmic}

\usepackage{xcolor}

\usepackage{upgreek}

\usepackage[utf8]{inputenc}

\journal{Expert Systems with Applications}

\biboptions{authoryear}

\begin{document}
\begin{frontmatter}


\title{Multivariate Forecasting of Bitcoin Volatility with Gradient Boosting: Deterministic, Probabilistic, and Feature Importance Perspectives}

\author[label1]{Grzegorz Dudek \corref{cor1}}
\ead{grzegorz.dudek@pcz.pl}

\author[label2]{Mateusz Kasprzyk}
\ead{aeris451@outlook.jp}

\author[label1]{Paweł Pełka}
\ead{pawel.pelka@pcz.pl}

\cortext[cor1]{Corresponding author.}
\address[label1]{Faculty of Electrical Engineering, Czestochowa University of Technology, Al. Armii Krajowej 17, Czestochowa, 42-201, Poland}
\address[label2]{Faculty of Computer Science and Artificial Intelligence, Czestochowa University of Technology, ul. Dabrowskiego 73, Czestochowa, 42-201, Poland}

\begin{abstract}
This study investigates the application of the Light Gradient Boosting Machine (LGBM) model for both deterministic and probabilistic forecasting of Bitcoin realized volatility. Utilizing a comprehensive set of 69 predictors -- encompassing market, behavioral, and macroeconomic indicators -- we evaluate the performance of LGBM-based models and compare them with both econometric and machine learning baselines. For probabilistic forecasting, we explore two quantile-based approaches: direct quantile regression using the pinball loss function, and a residual simulation method that transforms point forecasts into predictive distributions. To identify the main drivers of volatility, we employ gain-based and permutation feature importance techniques, consistently highlighting the significance of trading volume, lagged volatility measures, investor attention, and market capitalization. The results demonstrate that LGBM models effectively capture the nonlinear and high-variance characteristics of cryptocurrency markets while providing interpretable insights into the underlying volatility dynamics.
\end{abstract}


\end{frontmatter}

\section{Introduction}

The Bitcoin market is one of the most prominent and dynamic segments of the global financial system. As the leading cryptocurrency by market capitalization, Bitcoin plays a central role in the development of decentralized finance and is increasingly recognized as a speculative asset, a digital store of value, and even a portfolio diversifier. However, compared to traditional financial markets -- such as equities or foreign exchange -- Bitcoin is characterized by significantly higher volatility, lower liquidity, and greater sensitivity to market sentiment, regulatory news, and technological developments.

Due to this pronounced volatility, accurate forecasting of Bitcoin's variance is of substantial importance. Deterministic forecasting provides point estimates that are essential for portfolio optimization, risk management, and algorithmic trading. However, given the uncertainty and nonlinearity inherent in cryptocurrency markets, probabilistic forecasting offers a more comprehensive view by modeling the entire conditional distribution of volatility. This allows market participants to quantify uncertainty, construct prediction intervals, and make more informed decisions under risk. Therefore, combining both deterministic and probabilistic approaches enhances the robustness and applicability of volatility forecasting in the Bitcoin market.

In addition to accurate forecasting, the identification of key drivers of Bitcoin volatility is crucial for a deeper understanding of market dynamics. Unlike traditional assets, Bitcoin is influenced by a wide range of heterogeneous factors, including macroeconomic indicators, equity market behavior (e.g., S\&P 500), trading volume, investor sentiment, blockchain activity, and regulatory developments. Recognizing which variables contribute most to volatility allows researchers and practitioners to interpret forecast results more meaningfully and to design more effective trading, hedging, and risk control strategies.

\subsection{Related Work}

\subsubsection{Deterministic Forecasting of Cryptocurrency Volatility}

Forecasting cryptocurrency volatility, traditionally approached through deterministic (point) forecasts, attracted substantial academic interest, with models broadly divided into econometric and machine learning (ML) approaches. Traditional econometric frameworks -- such as ARCH/GARCH models -- remain in use but face criticism for their reliance on stationarity assumptions and inability to handle volatility jumps. For instance, \citep{Koc24} highlights that such models often generate implausible “exploding” processes and fail to accommodate infrequent but significant spikes \citep{charles2019volatility}.

Heterogeneous Autoregressive (HAR) models are regarded as more suitable alternatives, as they are estimated using realized variance (RV) rather than squared returns, as is the case with GARCH-type models. This superiority is evidenced in \citep{Ber22}, which compares various GARCH models with different error distributions (GARCH, EGARCH, GJR-GARCH, IGARCH, MSGARCH, and APARCH) against HAR. Extensions such as HAR-RV with jump components (HARRVJ) aim to account for structural breaks \citep{shen2020forecasting}, yet they still rely on the assumption of linear dynamics.

By contrast, ML and deep learning models do not rely on these assumptions. Instead, they can capture complex nonlinear relationships, structural breaks, and extreme volatility spikes more effectively. They achieve this through flexible architectures, powerful time-series decomposition techniques, sophisticated optimization, and ensembling strategies.
Recent high-impact examples include:
\citep{Her22}, where a Synthesizer Transformer trained on on‑chain metrics and "whale-alert" indicators demonstrated superior performance in forecasting volatility spikes;  
\citep{Wan23} showed that random forests and LSTM optimized via genetic algorithms and artificial bee colony outperform GARCH models for crypto volatility, with SHAP analysis identifying lagged volatility and moving-average features as key predictors;  
\citep{Zho25} introduced an Evolving Multiscale Graph Neural Network (EMGNN) that captures cross‑asset interactions, outperforming econometric and ML baselines.

Comparative studies further clarify this landscape: \citep{pratas2023forecasting} found that LSTM models better capture extreme volatility spikes than MLP or RNN, albeit at higher computational cost; \citep{Hua24} demonstrated that LSTM and hybrid CNN-LSTM models outperform both GARCH and HAR models across horizons ranging from 1 day to 2 months, with CNN-LSTM excelling in short-term prediction; and \citep{Dud24} compared econometric (HAR, ARFIMA, GARCH) and ML (LASSO, SVR, MLP, RF, LSTM) models in forecasting daily and weekly RV, finding no single model universally superior and that simple linear methods can rival more complex alternatives.

\subsubsection{Probabilistic Forecasting of Cryptocurrency Volatility}

Probabilistic forecasting provides full predictive distributions rather than single-point estimates, enabling the quantification of forecast uncertainty through prediction intervals or quantiles. Despite these advantages, probabilistic approaches remain underexplored in the context of cryptocurrency volatility.

One example is \citep{Gol24}, which proposed a model based on probabilistic gated recurrent units to generate predictive distributions. However, despite the model's probabilistic formulation, the evaluation was limited to point forecast accuracy, with no assessment of distributional quality. Another approach, introduced in \citep{Hon24}, leverages a variational autoencoder framework for multivariate distributional forecasting. This method directly estimates the cumulative distribution function of future conditional distributions, allowing for probabilistic forecasting by simulating synthetic future time series paths.

A more recent contribution by \citep{dudek2025probabilistic} presents probabilistic methods that build on point forecasts from a diverse set of statistical and ML-based models to estimate conditional quantiles of cryptocurrency RV. The proposed approaches include Quantile Linear Regression and Quantile Regression Forest as ensemble learners within a stacking framework, as well as a Quantile Estimation through Residual Simulation method.

\subsubsection{Identification of Bitcoin Volatility Drivers}

Identifying the key determinants of cryptocurrency volatility is essential for developing interpretable and effective forecasting models. Recent studies have employed a range of statistical and ML techniques to quantify the relative importance of potential predictors \citep{Wan23a,Kyr21}. Commonly analyzed factors include financial market indicators (e.g., stock indices, exchange rates, oil and gold prices), measures of investor attention and sentiment, macroeconomic uncertainty indices, and trading volume \citep{fiszeder2025bitcoin}.

In a regression-based study, \citep{Lyo20} used augmented HAR and non-crossing quantile HAR models to examine the determinants of Bitcoin volatility. The results indicated that volatility and its jump component are primarily driven by Bitcoin-specific risk factors, particularly regulatory news and hacking incidents involving cryptocurrency exchanges. Macroeconomic announcements generally had little effect, except for forward-looking indicators such as consumer confidence, suggesting a weak connection between Bitcoin volatility and broader economic conditions.

Paper \citep{Bou23} proposed CoMForE, a multimodal AdaBoost-LSTM ensemble integrating sentiment signals (from tweets and search queries), hash rate, and trading metrics. Their feature importance analysis highlighted public interest and blockchain-level metrics as dominant volatility drivers. Similarly, \citep{Her22} applied Explainable AI (XAI) methods to a Synthesizer Transformer model trained on CryptoQuant and whale-alert data, confirming the predictive relevance of both on-chain activity and large-scale transaction alerts.

In a comprehensive study, \citep{fiszeder2025bitcoin} combined statistical and ML techniques -- Bayesian Model Averaging, LASSO, and random forests -- to identify key predictors from a set of 62 variables, including market behavior, Google Trends, financial indices, and macroeconomic indicators. The most influential drivers included lagged daily, weekly, and monthly RV of Bitcoin, trading volume, and search interest in Bitcoin, with their importance found to be time-varying.



\subsection{Motivation and Contributions}

Volatility forecasting in the cryptocurrency market, particularly for Bitcoin, remains a challenging yet crucial task due to the asset's extreme price variability, sensitivity to sentiment and speculative behavior, and the absence of intrinsic valuation anchors. While traditional econometric models like GARCH and HAR have provided valuable insights, their performance is limited by assumptions of linearity and stationarity, which are often violated in high-frequency crypto markets. Meanwhile, recent advances in ML have opened new avenues for capturing complex, nonlinear dependencies and temporal structures in financial time series.

Despite the growing interest in ML-based volatility forecasting, the literature still lacks comprehensive studies that integrate state-of-the-art ML methods with probabilistic forecasting, particularly in the context of cryptocurrency markets. Most existing models focus on point forecasts, providing limited information about the uncertainty associated with predictions. In parallel, few studies systematically investigate the drivers of Bitcoin volatility and their relative importance in shaping forecast outcomes. 
The ability to quantify and rank predictor relevance provides transparency in ML-based models, often criticized as "black boxes", and offers valuable economic insights into the underlying mechanisms governing volatility in the cryptocurrency domain.

This study addresses these research gaps by proposing a novel framework for both deterministic and probabilistic multivariate forecasting of Bitcoin RV. We employ Light Gradient Boosting Machine (LGBM) known for its efficiency and high predictive accuracy, and explore two complementary quantile forecasting approaches. Importantly, our framework also provides feature importance estimates, enabling the identification of key volatility drivers across models. 

Our key contributions include:

\begin{enumerate}
    \item Proposing and rigorously evaluating the LGBM for multivariate Bitcoin volatility forecasting, addressing both deterministic (point) and probabilistic (quantile) predictions. We explore two quantile forecasting approaches: direct quantile regression with pinball loss and a residual simulation method that transforms point forecasts into probabilistic ones.
    \item Conducting an in-depth analysis of Bitcoin volatility drivers by using both gain-based feature importance and permutation feature importance on an extensive set of 69 (or 138 with shock indicators) predictors. This analysis also includes examining the impact of newly introduced shock indicators and the temporal dynamics of feature relevance.
    \item Performing a robust comparative study with multiple forecast accuracy metrics against established econometric and ML baseline models, consistently demonstrating the superior performance of LGBM-based variants in both deterministic and probabilistic forecasting scenarios.
    \item Providing insights into the importance of LGBM hyperparameters by applying functional ANOVA in conjunction with Bayesian optimization, thereby enabling a deeper understanding of their impact on model performance and guiding more effective hyperparameter tuning strategies in future applications.
\end{enumerate}

The rest of the paper is organized as follows. Section 2 details the comprehensive dataset, including the target variable (Bitcoin RV) and an extensive set of 69 predictors, further augmented by shock indicators. Section 3 formally states the problem for both deterministic and probabilistic Bitcoin volatility forecasting. Section 4 provides an in-depth explanation of the LGBM framework, covering its algorithm, loss functions, regularization, hyperparameters, quantile forecasting, and feature importance estimation techniques. Section 5 presents the experimental study, detailing preprocessing, training, optimization, baseline models, evaluation metrics, main results for both forecasting types, and analysis of predictor and hyperparameter importance. Finally, Section 6 concludes the paper.

\section{Data}
\label{Data}

Our target variable is the daily realized variance of the BTC/USD exchange rate (referred to as Bitcoin RV or RVBTC), defined as:

\begin{equation}
RV_{d,t} = \sum_{k=1}^{K} r_{k,t}^2, \qquad r_{k,t} = \ln P_{k,t} - \ln P_{k-1,t}
\end{equation}
where $K$ denotes the number of intraday return observations per day (e.g. $K=288$ for 5-minute returns), $r_{k,t}$ is the $k$-th intraday return on day $t$, and $P_{k,t}$ is the Bitcoin price at the $k$-th observation within day $t$. 

This target variable is forecasted using the data described in the following sections.

\subsection{Original Dataset}

The dataset is available at the following address: \url{https://doi.org/10.18150/SJHAHR}. A comprehensive description of the data can be found both there and in \citep{fiszeder2025bitcoin}. The data come from various sources, which are detailed in Table 2 of \citep{fiszeder2025bitcoin}. 
The dataset spans the period from 2 August 2017 to 31 March 2023 and consists of daily time series recorded from Monday to Friday, corresponding to the operating days of financial markets. The total length of the time series is 1,478 observations.

The data includes the logarithms of Bitcoin RV, denoted as {ln\_RVBTC}, as well as the logarithms of first lags of daily, weekly, and monthly realized variances, denoted as {ln\_RVBTCd}, {ln\_RVBTCw}, and {ln\_RVBTCm}, respectively. In addition, it contains exogenous variables derived from various data sources, categorized into three main groups:
\begin{enumerate}
    \item BTC-specific factors,
    \item financial markets, and
    \item market and policy uncertainty.
\end{enumerate}

The first group includes factors such as trading volume, number of daily transactions, average block size per day, market capitalization, average daily hash rate, number of unique addresses per day, and Google search trends for Bitcoin.

The second group, financial markets, is the most extensive. It includes major stock indices (S\&P 500, Nasdaq Composite, Euro STOXX 50, FTSE 100, Nikkei 225, Shanghai Composite), exchange rates (EUR/USD, JPY/USD, CNY/USD), the nominal broad U.S. dollar index, U.S. 2-year and 10-year bond yields, and various commodities (NYMEX light sweet WTI crude oil, gold spot price, NYMEX Henry Hub natural gas, and the Bloomberg Commodity Index).

The third group captures market and policy uncertainty and includes the CBOE Volatility Index, EURO STOXX 50 Volatility Index, CBOE Gold ETF Volatility Index, a newspaper-based economic policy uncertainty index, a Twitter-based market uncertainty index, a Twitter-based economic uncertainty index, a newspaper-based geopolitical risk index, a financial-variable-based risk aversion index, a financial-variable-based uncertainty index, and an infectious disease equity market volatility tracker.

Data in the first and third groups are transformed in two ways: by taking logarithms and first differences of logarithms. These transformed variables are denoted with the prefixes {ln\_} and {$\Updelta$\_}, respectively.
For each financial series in the second group, two derived variables are calculated: logarithmic returns and the Parkinson variance estimator, labeled with the prefixes {r\_} and {v\_}, respectively.

\begin{figure}[t]
\centering
\includegraphics[height=0.22\textwidth]{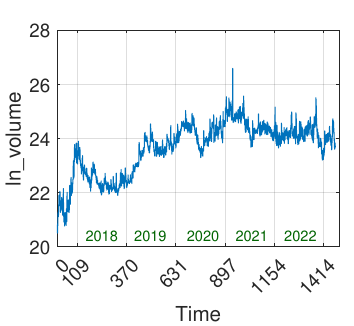}
\includegraphics[height=0.22\textwidth]{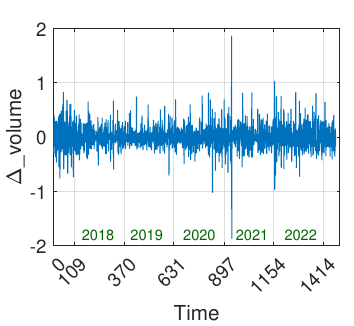}
\includegraphics[height=0.22\textwidth]{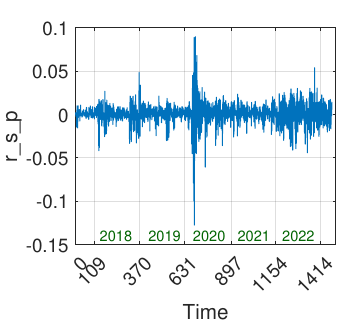}
\includegraphics[height=0.22\textwidth]{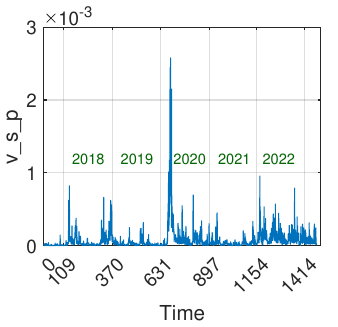}
\caption{Examples of exogenous predictors of different types (trading volume and S\&P 500 variables).}
\label{figTS}
\end{figure}

\begin{figure}[t]
\centering
\includegraphics[height=0.22\textwidth]{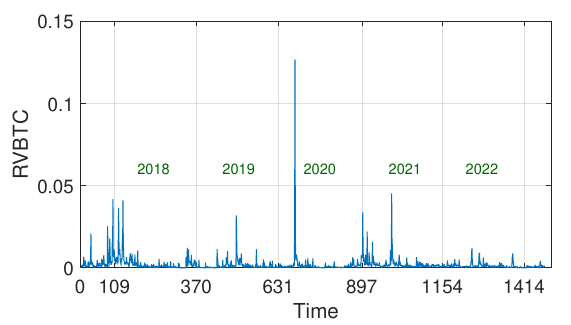}
\includegraphics[height=0.22\textwidth]{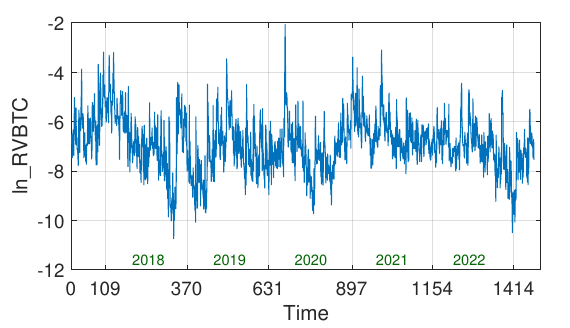}
\caption{Target variable: Bitcoin RV (RVBTC) and its logarithm (ln\_RVBTC).}
\label{figTSa}
\end{figure}

The variables prefixed with ln\_, $\Updelta$\_, r\_, and v\_ represent the set of predictors, comprising a total of 69 variables. These four categories are illustrated in Fig. \ref{figTS}, while Fig. \ref{figTSa} presents the target variable -- Bitcoin RV and its logarithm. Note that RVBTC exhibits multiple spikes and follows a distribution that is approximately exponential. These characteristics pose a challenge for forecasting models. To address this, models often operate on the log-transformed target variable, which follows a distribution closer to normal and reduces the influence of outliers.

\subsection{Shock Indicators}
\label{secSI}

For each original predictor $x_j$, we define a corresponding shock indicator as a binary variable, $x'_j$. A shock is identified when the absolute value of $x_j$ exceeds a threshold proportional to its standard deviation:

\begin{equation}
x'_{j,t} = \begin{cases}
1, & \text{if } |x_{j,t}| > \gamma \sigma_j, \\
0, & \text{otherwise}
\end{cases}
\end{equation}
where $t=1, ..., N$, with $N$ denoting the number of samples, $j=1, ..., n$, with $n$ representing the number of original predictors, $\gamma$ is a threshold parameter (treated as a tunable hyperparameter), and $\sigma_j$ denotes the standard deviation of the predictor $x_j$.

\begin{figure}[t]
\centering
\includegraphics[width=0.50\textwidth]{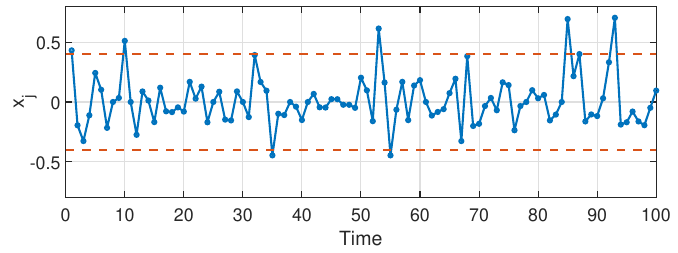}
\includegraphics[width=0.50\textwidth]{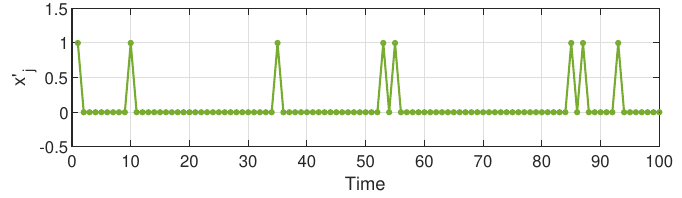}
\caption{Example of an original predictor and its corresponding shock indicator.}
\label{fig1}
\end{figure}

Note that the threshold is dynamically scaled according to the variability of each predictor. An example illustrating both the original predictor $x_j$ and its corresponding shock indicator $x'_j$ is presented in Fig.~\ref{fig1}.

Shock indicators are introduced to provide the model with additional information about extreme values of the predictors. The aim is to enhance the model's sensitivity to rare but impactful deviations.
In the experimental section, we examine two model variants: one without and one with the inclusion of shock indicators as additional predictors.


\section{Problem Statement} \label{PS}

\subsection{Deterministic Forecasting}
\label{DF}

Let $\mathbf{x}_i \in \mathbb{R}^n$ denote the predictor vector for observation $i$, composed of lagged variables available at time $t-1$ (i.e. ln\_RVBTCd($t-1$), ln\_volume($t-1$), r\_s\_p($t-1$), ...). These include either 69 original predictors or 138 predictors when incorporating the shock indicators described in Section~\ref{secSI}. Let $y_i \in \mathbb{R}$ denote the corresponding target variable to be forecasted at time $t$ -- in our case, {ln\_RVBTC}($t$).

The objective of deterministic forecasting is to learn a function $f: \mathbb{R}^n \rightarrow \mathbb{R}$ that maps the predictor vector $\mathbf{x}_i$ to a point estimate $\hat{y}_i$ of the target variable:

\begin{equation}
    \hat{y}_i = f(\mathbf{x}_i)
\end{equation}

The function $f$ is estimated using historical data $\Phi = \{ (\mathbf{x}_i, y_i) \}_{i=1}^{N}$, where $N$ denotes the size of the training set. The model is trained by minimizing a suitable loss function. In our study, $f$ is estimated using the LGBM algorithm, with the loss function defined in Section~\ref{LossDe}.

\subsection{Quantile Forecasting}

In the quantile forecasting setting, the objective is to estimate the conditional quantiles of the target variable $y_i$ at time $t$, given the predictor vector $\mathbf{x}_i$.

Instead of learning a single function as in deterministic forecasting, we aim to estimate a family of functions $\{f_q: \mathbb{R}^n \rightarrow \mathbb{R} \}_{q \in \Pi}$, where each $f_q(\mathbf{x}_i)$ returns the predicted $q$-th conditional quantile of $y_i$, and $\Pi$ is a predefined set of quantile levels:

\begin{equation}
    \hat{y}_{q,i} = f_q(\mathbf{x}_i), \quad \text{for each } q \in \Pi,
\end{equation}
where $\hat{y}_{q,i}$ denotes the predicted $q$-th quantile of $y_i$.

In this study, we consider 99 quantile levels: $\Pi = \{0.01, 0.02, \ldots, 0.99\}$. The functions $f_q$ are estimated using LGBM models trained on the set $\Phi$ by minimizing the pinball loss (also known as the quantile loss), as defined in Section~\ref{LossPr}. This formulation enables modeling of the entire conditional distribution of the target variable, supporting probabilistic forecasting and uncertainty quantification.

To avoid training a separate model for each quantile level, which can be computationally expensive, Section~\ref{SecQF} introduces a simplified approach for quantile estimation based on residual simulation using historical data.

\section{Light Gradient Boosting Framework}
\label{sec:LGBM}

Gradient Boosting Machines (GBMs) have emerged as powerful and versatile machine learning algorithms, achieving state-of-the-art results across a wide range of tasks, including classification, regression, and ranking \citep{friedman2001greedy,chen2016xgboost}. At their core, GBMs are ensemble methods that sequentially build an additive model by fitting new base learners (typically decision trees) to the residual errors of the previous learners. This iterative process allows GBMs to capture complex nonlinear relationships in the data.

While traditional GBM implementations are highly effective, they can be computationally expensive, particularly when dealing with large datasets with a high number of features. This computational burden often limits their applicability in resource-constrained environments or real-time applications. To address these limitations, \citep{ke2017lightgbm} introduced the Light Gradient Boosting Machine, a highly efficient and scalable gradient boosting framework.

LGBM incorporates several key innovations that significantly improve its speed and memory efficiency without sacrificing accuracy. These core techniques include:

\begin{itemize}

    \item \textbf{Gradient-Based One-Side Sampling (GOSS)}: To accelerate training, GOSS retains instances with large gradients and randomly samples instances with small gradients. This technique focuses on the most informative data points, maintaining accuracy while reducing computation.
    
    \item \textbf{Histogram-Based Decision Tree Learning}: Instead of evaluating all possible split points, LGBM discretizes continuous feature values into discrete bins, constructing histograms to find optimal splits. 
    This approach significantly reduces computational complexity and memory usage. Additionally, it accounts for the data distribution and gradient statistics within each node, enabling the algorithm to efficiently identify near-optimal split points.

    \item \textbf{Leaf-Wise Tree Growth with Depth Limitation}: Unlike level-wise tree growth strategies, LGBM grows trees leaf-wise, choosing the leaf with the maximum loss reduction to split. This method can lead to more complex, deeper trees and potentially better accuracy, while the depth limitation prevents overfitting.

    \item \textbf{Exclusive Feature Bundling (EFB)}: LGBM bundles mutually exclusive features (features that rarely take non-zero values simultaneously) into a single feature, reducing the number of features and further improving efficiency.

    \item \textbf{Support for Parallel and GPU Learning}: LGBM supports parallel learning and GPU acceleration, enabling faster training on large datasets.

\end{itemize}

The combination of these innovative techniques makes LGBM a highly competitive and widely adopted gradient boosting framework, particularly for large-scale machine learning tasks where speed and memory efficiency are critical. Its performance has been demonstrated across various domains, often outperforming other gradient boosting implementations in terms of both speed and accuracy \citep{ke2017lightgbm,zhang2020comparative}. In the financial domain, LGBM has been successfully applied to a range of problems, including credit scoring, fraud detection, and risk assessment \citep{sym15040870,LI2025200514,WANG2022259}.

\subsection{Algorithm}

The LGBM algorithm operates through the following iterative procedure:

\begin{enumerate}
    \item \textbf{Boosting Process:} 
    LGBM constructs an ensemble model in an additive manner, training one decision tree at a time. Each subsequent tree is designed to correct the errors (pseudo-residuals) made by the ensemble of previously trained trees.

    \item \textbf{Prediction up to Iteration $m-1$:}
    Before training the $m$-th tree, we already have a model $F_{m-1}(\textbf{x})$ consisting of $m-1$ learned trees. For each training instance $\textbf{x}_i$, this model generates a prediction $\hat{y}_{i,m-1} = F_{m-1}(\textbf{x}_i)$.

    \item \textbf{Calculating Pseudo-Residuals:} For each training instance $(\textbf{x}_i, y_i)$, the pseudo-residuals are calculated with respect to the loss function $L(y, F_{m-1}(\textbf{x}))$. The pseudo-residuals are essentially the negative gradient of the loss function with respect to the prediction of the model $F_{m-1}(\textbf{x}_i)$ for each instance:

\begin{equation} \label{eqg}
    g_{i,m} = -\frac{\partial L(y_i, F_{m-1}(\textbf{x}_i))}{\partial F_{m-1}(\textbf{x}_i)}, \quad i=1, ..,N
\end{equation}

    These pseudo-residuals represent the direction and magnitude by which the model's predictions should be adjusted to minimize the loss function.
    

    \item \textbf{Training the $m$-th Tree:} The $m$-th tree $h_m(\textbf{x})$ is trained to predict these pseudo-residuals $g_{i,m}$ based on features $\textbf{x}_i$. In other words, the goal of the $m$-th tree is to model the errors made by the previous ensemble $F_{m-1}$.

    \item \textbf{Updating the Model:} After training the $m$-th tree $h_m(\textbf{x})$, the model is updated by adding the new tree, scaled by the learning rate $\eta$:

\begin{equation} \label{eqFm}
    F_m(\textbf{x}) = F_{m-1}(\textbf{x}) + \eta h_m(\textbf{x})
\end{equation}

    This update aims to refine the model's predictions by incorporating the corrections suggested by $h_m$.
\end{enumerate}

\begin{figure}[t]
\centering
\includegraphics[width=1\textwidth]{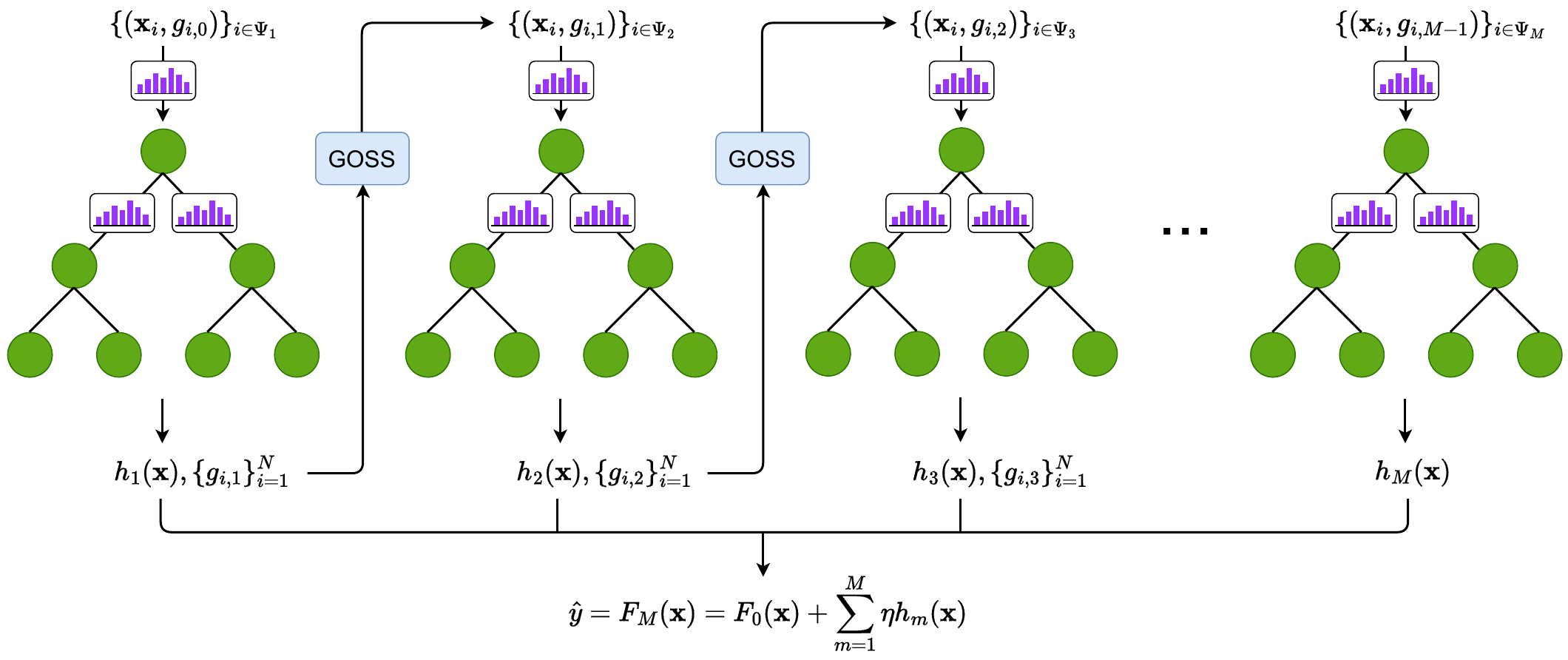}
\caption{LGBM model.}
\label{fig-1}
\end{figure}

The LGBM algorithm is illustrated in Fig.~\ref{fig-1}. Here, $F_0(\mathbf{x})$ represents the initial prediction (base model), typically set as the average of the target variable $y$ from the training data. This initial model provides the starting pseudo-residuals $g_{i,0}$, which serve as the errors for the first tree to learn.

GOSS dynamically selects training samples for each tree based on the magnitude of their pseudo-residuals. As a result, each tree is trained on a distinct subset of the data, $\Psi_m$. 
Alternatively to GOSS, GBDT (Gradient Boosting Decision Tree) can be used, which relies on all training samples at each boosting iteration (for every tree). It may be more stable but computationally heavier.

Crucially, the target values for the $m$-th tree are the pseudo-residuals $g_{i,m-1}$, not the original labels $y_i$. This approach allows the model to focus on minimizing the loss function iteratively by addressing the specific errors made in previous iterations.

Fig.~\ref{fig-1} also highlights the histogram-based splitting mechanism. During tree construction, histograms are dynamically built for each feature at every node, discretizing continuous values into bins. 

In addition to the gradient, LGBM also utilizes the Hessian (second derivative of the loss function) to more accurately approximate the local curvature of the loss surface. This second-order information facilitates the calculation of optimal leaf weights and split gains, enhancing training stability, convergence speed, and overall predictive performance.

In this study, we apply the LGBM algorithm, defined as described above, for both deterministic and probabilistic forecasting of Bitcoin volatility. The distinction between these approaches lies in the choice of the loss function used during training.

\subsection{Loss Function for Deterministic Forecasts}
\label{LossDe}

For point forecast we minimize the fair loss function, which is defined as: 

\begin{equation} \label{eqF}
    L(y, \hat{y}) = c^2 \left( \frac{|y - \hat{y}|}{c} - \ln\left(1 + \frac{|y - \hat{y}|}{c}\right) \right)
\end{equation}
where $c$ is a scaling parameter that controls the transition point between different behaviors.

\begin{figure}[t]
\centering
\includegraphics[width=0.40\textwidth]{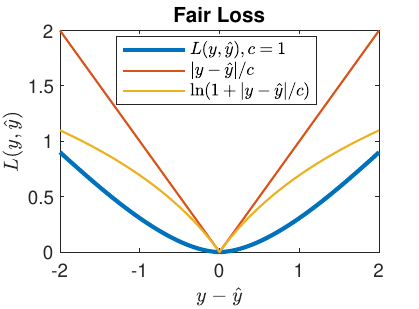}
\includegraphics[width=0.40\textwidth]{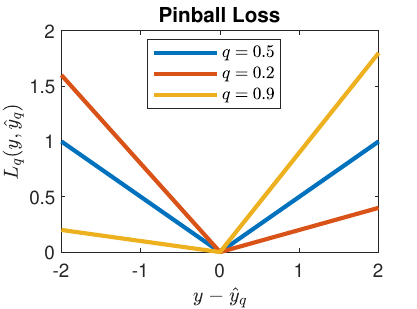}
\caption{Loss functions for point (left) and quantile (right) forecasting.}
\label{figLoss}
\end{figure}

Fair loss is a robust regression loss function that adaptively adjusts its behavior based on the magnitude of prediction errors. For small residuals, it behaves similarly to the Mean Squared Error (MSE), exhibiting a quadratic form that facilitates precise fine-tuning of predictions. For larger residuals, however, it transitions to a linear form, akin to the Mean Absolute Error (MAE), thereby reducing sensitivity to outliers. This hybrid structure enables fair loss to overcome key limitations of both MSE and MAE: while MSE disproportionately penalizes large errors, leading to instability in the presence of outliers, MAE suffers from non-differentiability at zero, which can hinder gradient-based optimization. By combining the advantages of both, fair loss offers improved robustness and smoother convergence properties. Its functional form is illustrated in Fig. \ref{figLoss}.

The parameter $c$ controls the sensitivity to outliers: smaller values make the loss function more robust to extreme deviations, while larger values increase sensitivity to all errors. The flexibility of fair loss makes it particularly well-suited for real-world applications, such as financial forecasting, where models must balance precision for typical observations with resilience to rare but significant anomalies.

\subsection{Loss Function for Quantile Forecasts}
\label{LossPr}



The pinball loss function is a fundamental component of quantile regression, used to assess how well a model estimates a specific quantile $y_q$ of the target variable $y$. For a given quantile level $q \in (0, 1)$, the pinball loss is defined as:

\begin{equation}
L_q(y, \hat{y}_q) = \begin{cases}
q(y - \hat{y}_q) & \text{if } y \geq \hat{y}_q \\
(q-1)(y - \hat{y}_q) & \text{if } y < \hat{y}_q
\end{cases}
\label{eqrf}
\end{equation}

This loss function is inherently asymmetric, allowing it to penalize under-predictions and over-predictions differently depending on the quantile level $q$.
Specifically, when $q > 0.5$, under-predictions are penalized more heavily than over-predictions; conversely, when $q < 0.5$, the loss assigns a greater penalty to over-predictions. At $q = 0.5$, the pinball loss reduces to MAE, applying equal penalty to both types of errors. This behavior is illustrated in Fig. \ref{figLoss}, where the slope discontinuity at $y-\hat{y}_q=0$ highlights the function’s piecewise linear and asymmetric nature.

\subsection{Regularization}

In LGBM, regularization plays a critical role in controlling model complexity and improving generalization performance. The two primary forms of regularization employed are: $L_1$ (Lasso), 
which promotes sparsity by shrinking less important leaf weights to zero,
and $L_2$ (Ridge) regularization, which prevents overfitting by penalizing large leaf weight values. Both methods are applied directly to the leaf weights $\omega_j$, which represent the predicted values at the terminal nodes of the decision trees.

The objective function that LGBM optimizes consists of a loss function and a regularization term defined as:

\begin{equation}
\Omega(\mathbf{\omega}) = \alpha \sum_{j=1}^{T} |\omega_j| + \frac{\lambda}{2} \sum_{j=1}^{T} \omega_j^2 + \gamma T
\label{eqreg}
\end{equation}
where $T$ is the total number of leaves, $\omega_j$ denotes the weight of the $j$-th leaf, $\alpha$ and $\lambda$ are the coefficients for $L_1$ and $L_2$ regularization, respectively, and $\gamma$ 
is the complexity penalty, representing the minimum loss reduction required to make a further split.

\subsection{Hyperparameters}

In our implementation of LGBM for Bitcoin RV forecasting, we use the following hyperparameters:

\begin{itemize}
\item \texttt{n\_estimators}, $M$ -- number of boosting rounds (i.e., number of trees in the model),
\item \texttt{max\_depth} -- maximum depth of each tree (limits tree complexity to reduce overfitting),
\item \texttt{min\_data\_in\_leaf} -- minimum number of samples required in a leaf (controls model complexity and prevents overfitting),
\item \texttt{learning\_rate}, $\eta$ -- learning rate that scales the contribution of each tree (smaller values lead to slower but potentially more accurate learning),
\item \texttt{reg\_alpha}, $\alpha$ -- $L_1$ regularization term on leaf weights (Lasso regularization; encourages sparsity),
\item \texttt{reg\_lambda}, $\lambda$ -- $L_2$ regularization term on leaf weights (Ridge regularization; penalizes large weights),
\item \texttt{feature\_fraction} -- fraction of features randomly selected for building each tree (reduces overfitting, similar to random forest),
\item \texttt{bagging\_fraction} -- fraction of samples used to train each tree (used for row subsampling in GBDT; improves generalization),
\item \texttt{bagging\_freq} -- frequency of bagging in GBDT (apply bagging every \texttt{bagging\_freq} iteration),
\item \texttt{max\_bin} -- maximum number of bins used to discretize continuous features (affects accuracy vs. speed trade-off).
\end{itemize}

\subsection{Quantile Forecasting}
\label{SecQF}

We employ two approaches for quantile forecasting using LGBM. The first approach, as described above, utilizes LGBM with the pinball loss function. This method requires training a separate model for each quantile level individually. In the experimental section, this approach is denoted as LGBM-pinball.

The second approach, known as Quantile Estimation through Residual Simulation (QRS), is a post-processing technique designed to convert point forecasts into probabilistic forecasts by leveraging the empirical distribution of historical forecast errors. The QRS procedure unfolds as follows. 
For each test sample $\tau$, the model is trained on historical data to generate point predictions. 
Forecast residuals are then computed over the training period as the difference between the predicted and observed values: $e_t = y_t - \hat{y}_t, t \in \Phi_\tau$, where $\Phi_\tau$ is a training set for $\tau$.

From these historical residuals, $\Theta_{\tau} = \{e_t\}_{t \in \Phi_{\tau}}$, an empirical quantile distribution is derived. These quantiles are subsequently added to the point forecast for time $\tau$ to produce probabilistic predictions.

Mathematically, for a given point forecast $\hat{y}_{\tau}$, the corresponding quantile forecast at level $q$ is defined as: 

\begin{equation} \label{eqQRS}
    \hat{y}_{q,\tau} = \hat{y}_{\tau} + e_{q,\tau} \quad q \in \Pi
\end{equation}
where $\hat{y}_{\tau}$ and $\hat{y}_{q,\tau}$ represent the point forecast and predicted quantile at level $q$, respectively, $e_{q,\tau}$ is the $q$-th quantile of the residual distribution $\Theta_{\tau}$,
and $\Pi$ denotes the set of target quantile levels.

\subsection{Feature Importance Estimation}
\label{PrIm}

Feature importance refers to techniques used to quantify how much each input feature contributes to a model’s predictive performance.
In LGBM, feature importance can be estimated using two built-in methods: "split" (frequency-based) and "gain" (contribution-based). Both are derived from the structure of the trained decision trees but offer different perspectives on feature influence.

The split-based importance measures the number of times a feature is used to split the data across all trees. This method is useful for identifying which features are most frequently involved in the decision process but does not account for the quality or impact of those splits.

In contrast, the gain-based importance measures the total reduction in the loss function resulting from all splits involving a given feature. Features that produce greater improvements in model performance, i.e., higher reductions in the loss function, receive higher importance scores.

The gain-based method is typically preferred, as it provides a more nuanced assessment of feature relevance by incorporating both the frequency and the impact of splits. Therefore, in this study, we adopt gain-based feature importance to assess the contribution of predictors to Bitcoin volatility forecasting.

\subsubsection{Gain Feature Importance (GFI)} 
GFI is often normalized as a percentage of the total gain across all features, as defined below:

\begin{equation} \label{eqGFI}
{GFI}(j) = 100 \cdot \frac{\sum_{m=1}^{M} \sum_{k \in \Xi_m} \mathds{1}{\{\text{feature}(k) = j\}\cdot G_k}}{\sum_{j'=1}^{n} \sum_{m=1}^{M} \sum_{k \in \Xi_m} \mathds{1}\{\text{feature}(k) = j'\}\cdot G_k}
\end{equation}
where: $j$ is the feature of interest, 
$M$ is the total number of trees,
$n$ is the total number of features,
$\Xi_m$ is the set of all internal nodes in tree $m$,
$\mathds{1}{\{\cdot\}}$ is the indicator function,
$\text{feature}(k)$ is the feature used for splitting at node $k$, and
$G_k$ is the gain (loss reduction) achieved by the split at node $k$.

The gain from a split is determined by quantifying the reduction in the model's overall loss function achieved by that split, considering both the initial and resulting node gradients and Hessians, while also incorporating $L_1$ and $L_2$ regularization penalties and a minimum gain threshold.



\subsubsection{Permutation Feature Importance (PFI}





PFI is a model-agnostic technique that quantifies the importance of each predictor by evaluating the change in model performance when the values of a single feature are randomly shuffled. This permutation breaks the association between the feature and the target, thereby degrading the model's ability to make accurate predictions if the feature is informative.

The basic procedure for computing PFI is as follows:

\begin{enumerate}
  \item Fit the predictive model (here, LGBM) on the full training data and compute the baseline performance score (in our case, the coefficient of determination $R^2$).
  \item For each feature $j \in \{1, ..., n\}$:
  \begin{enumerate}
    \item Randomly permute the values of the $j$-th feature in the dataset.
    \item Compute the performance score $R^2_j$ on the modified dataset.
    \item Define the importance of feature $j$ as the difference between the baseline score and the permuted score: $pfi(j) = R^2 - R^2_j$
  \end{enumerate}
  \item Repeat the permutation multiple times (five iterations in our case) to obtain a robust estimate of importance, averaging the resulting scores.
\end{enumerate}

The final PFI score is expressed as a percentage:

\begin{equation} \label{eqPFI}
{PFI}(j) = 100 \cdot \frac{pfi(j)}{\sum_{j'=1}^{n} pfi(j')}
\end{equation}

This method has the advantage of directly assessing the marginal contribution of each feature to the model's predictive ability. In contrast to split- or gain-based importance measures, which rely on the internal structure of decision trees, PFI is independent of the model's internal mechanics and can capture interactions and nonlinear effects.

Note that GFI and PFI do not reflect to the intrinsic predictive value of a feature by itself but how important this feature is for a particular LGBM model.

\section{Experimental Study}

In this section, we evaluate the performance of the LGBM model in both deterministic and probabilistic multivariate forecasting of Bitcoin volatility. The analysis includes comparisons with baseline models, assessment of predictor importance, and sensitivity analysis of the model with respect to its hyperparameters.

{Although the logarithm of RV was used as the target variable during model training to address heteroskedasticity, all evaluation metrics (MAE, MSE, CRPS, etc.) were calculated after back-transforming predictions to the original RV scale.}

The following notation is used to denote the different model variants:

    \begin{itemize}
        \item \textbf{LGBM} -- deterministic forecasting model without shock indicators,
        \item \textbf{LGBM32} -- ensemble of 32 LGBM models,
        \item \textbf{LGBMs} -- deterministic forecasting model with shock indicators,
        \item \textbf{LGBM32s} -- ensemble of 32 LGBMs models, 
        \item \textbf{LGBM-pinball} -- probabilistic forecasting model trained with the pinball loss function,
        \item \textbf{QRS-LGBM} -- probabilistic forecasting model based on the Quantile Residual Simulation approach (see Section~\ref{SecQF}).
    \end{itemize}

\subsection{Preprocessing, Training and Optimization Setup}

The data period spans from August 1, 2017, to March 31, 2023. The test sets were defined as follows:

\begin{itemize}
    \item for deterministic forecasting, data from January 1, 2020, to March 31, 2023, were used (848 test samples),
    \item for probabilistic forecasting, we used data from the year 2021 (261 test samples), adopting the same test period as in the baseline models \citep{dudek2025probabilistic} to ensure a fair comparison.
\end{itemize}

For each day in the test period, a separate LGBM model was trained using preceding data. This results in either 848 or 261 training runs per model, depending on the forecasting task. This rolling-origin approach ensures that the model consistently utilizes the most recent information, based on the assumption that the most relevant patterns for forecasting are embedded in the immediate past. The forecasting horizon was set to $h=1$.

Feature preprocessing included the detection and replacement of outliers in the time series of input variables. We employed winsorization with thresholds set at the 0.01 and 0.99 quantiles to cap extreme values while preserving the overall distributional structure. Following outlier treatment, all features were standardized. {Both winsorization and standardization were parameterized using only the training sets to avoid information leakage. 
The same principle was applied to the shock variables, which were recalculated at each step as the training window expanded to define the training data range for subsequent test samples.}

Before the main training phase, the models were optimized using data from the period preceding the test data period. Preliminary experiments showed that GOSS did not yield better performance than the standard GBDT algorithm. Consequently, all further experiments were conducted using GBDT.

Our optimization methodology employed a Bayesian optimization with a three-fold cross-validation strategy to identify optimal hyperparameter configurations. To ensure robustness and reduce variance in hyperparameter selection, we conducted multiple independent validation runs using different random seeds (42, 2023, and 999). This multi-seed approach provides more stable optimization outcomes by averaging performance across different data partitions and random initializations.

The hyperparameter search was conducted using the Optuna framework with the Tree-structured Parzen Estimator (TPE) algorithm \citep{optuna_2019}. Each optimization run consisted of 300 trials. For the probabilistic LGBM-pinball model, optimization was performed specifically for the 0.5 quantile. The objective function minimized MAE.  

The optimal hyperparameter configuration is summarized in Table~\ref{tabLGBM}. The search grids were determined based on preliminary experiments.
Notably, all models employ shallow trees with \texttt{max\_depth} = 2, indicating that such tree structures were sufficient to capture the relevant patterns in the data. 
The moderate learning rates (ranging from 0.02 to 0.04), combined with 500 or 600 estimators, provide adequate model capacity while ensuring training stability.
Regularization strengths (\texttt{reg\_alpha} and \texttt{reg\_lambda}) differ substantially across models, reflecting varying needs for sparsity and weight shrinkage. For example, the LGBM model applies strong $L_1$ regularization, while LGBMs relies more heavily on $L_2$. Feature and bagging fractions also vary, with LGBM-pinball employing more aggressive subsampling (\texttt{bagging\_fraction} = 0.5) to enhance generalization. The \texttt{max\_bin} parameter is considerably reduced in LGBM-pinball, potentially to accelerate training or mitigate overfitting. The \texttt{shock\_threshold} parameter applies only to models with shock indicators (LGBMs) and was optimized to a value of 2.5. 
In our implementation, we did not optimize the complexity penalty term in \eqref{eqreg}, instead setting $\gamma = 0$, nor did we tune the scaling parameter in \eqref{eqF}, which was fixed at $c = 1$.

\begin{table}[t]
\centering
\setlength{\tabcolsep}{6pt}
\caption{LGBM optimal hyperparameter settings from Bayesian optimization.}
\label{tabLGBM}
\begin{tabular}{lcccc}
\toprule
Hyperparameter & Grid & LGBM & LGBMs & LGBM-pinball\\
\midrule
\texttt{n\_estimators}, $M$& {[50, ..., 600]}& 500& 500& 600\\
\texttt{max\_depth} & {[1, 2]} & 2& 2& 2\\
\texttt{min\_data\_in\_leaf}& {[1, ..., 4]} & 3& 2& 4\\
\texttt{learning\_rate}, $\eta$ & {[0.02, ..., 0.04]} & 0.0234 & 0.0203 & 0.0399 \\
\texttt{reg\_alpha}, $\alpha$  & $[10^{-5}, ..., 0.2]$ & 0.132& $1.51\cdot10^{-5}$ & $8.5\cdot10^{-4}$\\
\texttt{reg\_lambda}, $\lambda$& $[10^{-5}, ..., 0.2]$ & 0.002& 0.197& $1.13\cdot10^{-4}$ \\
\texttt{feature\_fraction}& [0.5, ..., 1.0] & 0.6& 0.8& 0.6\\
\texttt{bagging\_fraction} & {[0.5, ..., 1.0]} & 1.0& 0.7& 0.5\\
\texttt{bagging\_freq} & {[1, ..., 6]} & 3& 3& 2\\
\texttt{max\_bin}& {[32, ..., 255]}& 128& 224& 32\\
\texttt{shock\_threshold}& [1.5, ..., 3.0] & -- & 2.5 & --\\
\bottomrule
\end{tabular}
\end{table}


To ensure reproducibility and statistical validity, all stochastic components of the pipeline were governed by fixed random seeds. This includes data splitting procedures, model initialization, and training processes. The combination of multiple random seed validations and fixed reproducibility seeds provides both robust hyperparameter selection and consistent experimental results across different runs.

\subsection{Baseline Models}

The baseline models for deterministic forecasting are adopted from \citep{fiszeder2025bitcoin}, where the same forecasting problem as in this study was considered (see Section \ref{DF}). The following models are included:
\begin{itemize}
    \item \textbf{HAR} -- Heterogeneous AutoRegressive model,
    \item \textbf{BMA} -- Bayesian Model Averaging for Linear Regression Models (referred to as \texttt{BMA-X\_out\_st} in \citep{fiszeder2025bitcoin}),
    \item \textbf{LASSO} -- Least Absolute Shrinkage and Selection Operator model (referred to as \texttt{LASSO-X\_out\_st} in \citep{fiszeder2025bitcoin}),
    \item \textbf{RF} -- Random Forest (referred to as \texttt{RF-X\_out} in \citep{fiszeder2025bitcoin}).\\
\end{itemize}

The baseline models for probabilistic forecasting are adopted from \citep{dudek2025probabilistic}:
\begin{itemize}
    \item \textbf{QRS-HAR} -- QRS approach based on HAR predictions (referred to as \texttt{QRS-l, HAR-l} in \citep{dudek2025probabilistic}),
    \item \textbf{QRS-Ridge} -- QRS approach based on Ridge Regression predictions (referred to as \texttt{QRS-l, RR-l} in \citep{dudek2025probabilistic}),
    \item \textbf{QLR} -- Quantile Liner Regression,
    \item \textbf{QRF} -- Quantile Random Forest.
\end{itemize}

All probabilistic baseline models were trained without exogenous variables, using only predictors derived directly from the Bitcoin RV time series, namely ln\_RVBTCd, ln\_RVBTCw, and ln\_RVBTCm.
Each baseline model was carefully optimized and trained. Detailed descriptions of the optimization and training procedures are provided in \citep{fiszeder2025bitcoin,dudek2025probabilistic}.

\subsection{Evaluation Metrics}

To evaluate point forecasts, we use typical metrics based on forecast error defined as $E=y-\hat{y}$: Mean Absolute Error (MAE), Symmetric Mean Absolute Percentage Error (sMAPE), Mean Squared Error (MSE), and Coefficient of Determination ($R^2$).

To evaluate quantile forecasts, we use:
\begin{itemize}
    \item Continuous Ranked Probability Score (CRPS):
\begin{equation} \label{eqCRPS}
    \text{CRPS} = \frac{2}{|\Pi|}\sum_{q \in \Pi} L_{q}(y, \hat{y}_{q})
\end{equation}
where $\Pi = \{0.01, 0.02, \ldots, 0.99\}$ is the set of quantile levels,
$\hat{y}_{q}$ represents the predicted quantile at level $q$, and  
$L_{q}(y, \hat{y}_{q})$ is the pinball loss, as defined in \eqref{eqrf}.

    \item Relative Frequency (ReFr, calibration) \citep{Mar22}:
\begin{equation}
\text{ReFr}(q)=\frac{1}{N} \sum_{i=1}^N \mathds{1}{\{y_i \leq \hat{y}_{q,i}\}}
\label{eqrefr}
\end{equation}
where $N$ is the number of samples and $\mathds{1}{\{\cdot\}}$ is the indicator function.

The expected value of $\text{ReFr}(q)$ is the nominal probability level 
$q$. In other words, the predicted $q$-quantiles should exceed the realized values in $100q\%$ of cases, ensuring a ReFr equal to 
$q$. 

To assess the average deviation of $\text{ReFr}(q)$ from the expected 
$q$ across all $q \in \Pi$, we use the Mean Absolute ReFr Error (MARFE) \citep{dudek2024stacking}:
\begin{equation}
\text{MARFE}=\frac{1}{|\Pi|} \sum_{q \in \Pi} |\text{ReFr}(q)-q|
\label{eqmrf}
\end{equation}

    \item Winkler Score (WS) \citep{Bre21}:
\begin{equation}
\text{WS} =
\begin{cases}
(\hat{y}_{q_u} - \hat{y}_{q_l}) + \frac{2}{\alpha}(\hat{y}_{q_l} - y) & \text{if } y < \hat{y}_{q_l}\\
(\hat{y}_{q_u} - \hat{y}_{q_l}) & \text{if } \hat{y}_{q_l} \leq y \leq \hat{y}_{q_u}\\
(\hat{y}_{q_u} - \hat{y}_{q_l}) + \frac{2}{\alpha}(y-\hat{y}_{q_u}) & \text{if } y > \hat{y}_{q_u}\\
\end{cases}
\end{equation}
where $\hat{y}_{q_l}$ and $\hat{y}_{q_u}$ represent the predicted lower and upper quantiles defining the $100(1-\alpha)\%$ prediction interval (PI), with $\alpha=q_u-q_l$.

In this study, we evaluate 90\% PIs ($q_l=0.05$ and $q_u=0.95$) using mean WS:
\begin{equation}
\text{MWS} = \frac{1}{N} \sum_{i=1}^{N} \text{WS}(y_i, \hat{y}_{q_l,i}, \hat{y}_{q_u,i})
\label{eqmws}
\end{equation}

    \item The percentages of observed values falling within, below, and above the 90\% PI. These intuitive metrics allow for an easy comparison of results against the expected values -- in our case, 90\%, 5\%, and 5\%, respectively.

    \item MAE-Q and MSE-Q, metrics for evaluating point forecasts derived from quantile forecasts. In this case, we assume that the point forecast is the predicted median: $\text{MAE-Q} = \frac{1}{N} \sum_{i=1}^{N} | y_i - \hat{y}_{0.5,i}|$, $\text{MSE-Q} = \frac{1}{N} \sum_{i=1}^{N} (y_i - \hat{y}_{0.5,i})^2$. 

\end{itemize}

\subsection{Results for Deterministic Forecasting}

Table~\ref{tab1} compares the performance of baseline models with the proposed model in four variants: LGBM (without shock indicators), LGBM32 (an ensemble of 32 LGBM models), LGBMs (with shock indicators), and LGBMs32 (an ensemble of 32 LGBMs). As shown in the table, all LGBM-based variants consistently yield lower prediction errors and significantly higher $R^2$ values compared to the baseline models. As expected, the ensemble variants (LGBM32 and LGBMs32) outperform their single-model counterparts. 
The comparison between LGBM models with and without shock indicators is inconclusive: the model with shock indicators achieves better MAE and $R^2$, but exhibits slightly higher MSE.

\begin{table}[]
\caption{Performance metrics for deterministic forecasts.}
\label{tab1}
\setlength{\tabcolsep}{4pt}
\centering
\begin{tabular}{lcccc}    
\toprule
Model   & MAE& sMAPE & MSE   & R$^2$    \\
\midrule 
HAR& 9.98E-04$\pm$4.77E-03 & 47.3$\pm$36.9 & 2.37E-05$\pm$5.05E-04 & 0.128 \\
BMA& 1.02E-03$\pm$4.72E-03 & 48.7$\pm$35.5 & 2.33E-05$\pm$5.02E-04 & 0.143 \\
LASSO   & 8.60E-04$\pm$4.34E-03 & 39.7$\pm$30.2 & 1.95E-05$\pm$4.23E-04 & 0.283 \\
RF & 9.48E-04$\pm$4.72E-03 & 42.6$\pm$32.9 & 2.31E-05$\pm$4.95E-04 & 0.150 \\
LGBMs   & 8.22E-04$\pm$2.98E-03 & 38.3$\pm$29.2 & 9.58E-06$\pm$1.50E-04 & 0.648 \\
LGBMs32 & \textbf{8.15E-04$\pm$2.94E-03} &\textbf{38.1$\pm$29.0}& 9.31E-06$\pm$1.43E-04 & 0.658 \\
LGBM    & 8.30E-04$\pm$2.85E-03 & 38.2$\pm$29.5 & 8.87E-06$\pm$1.06E-04 & 0.674 \\
LGBM32  & 8.28E-04$\pm$2.80E-03 & \textbf{38.1$\pm$29.4} & \textbf{8.50E-06$\pm$9.38E-05} & \textbf{0.688} \\
\bottomrule
\end{tabular}
\end{table}

\begin{figure}[t]
\centering
\includegraphics[width=0.475\textwidth]{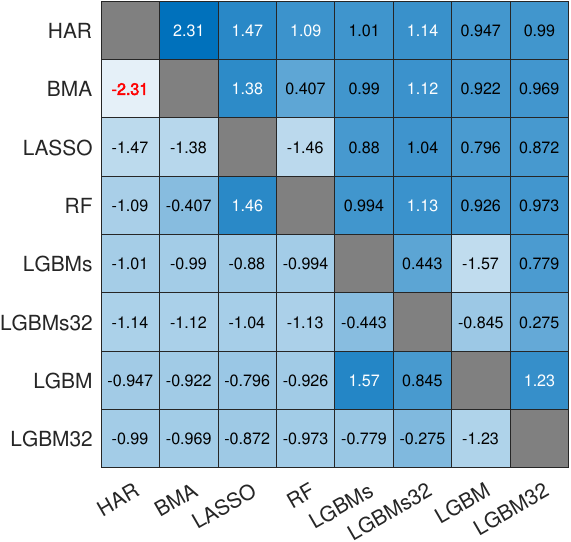}
\includegraphics[width=0.470\textwidth]{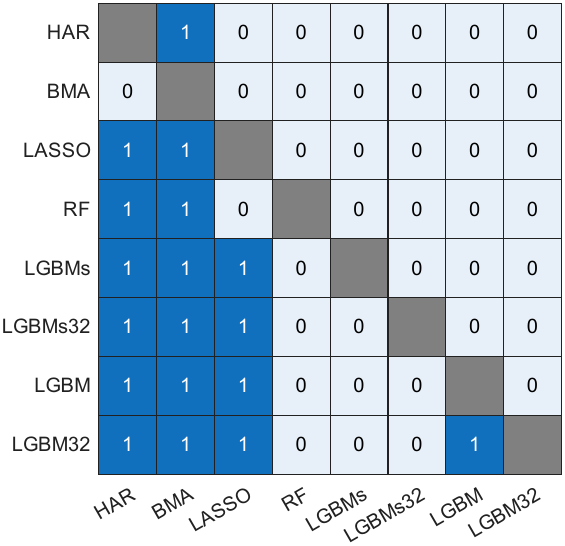}
\caption{{Diebold–Mariano statistics (left) and Wilcoxon signed-rank results (right) for deterministic forecasts. In the DM test, values below $-1.96$ indicate that the y-axis model has significantly lower forecast error (E) than the x-axis model at $\alpha = 0.05$. In the Wilcoxon test, a value of 1 indicates significantly lower errors for the y-axis model at the same significance level.}}
\label{figDM}
\end{figure}

To assess whether the differences in prediction errors are statistically significant, the Diebold-Mariano (DM) test \citep{Die95} was applied with a significance level of 
$\alpha=0.05$. This test evaluates whether the residuals ($E$) from two competing models differ significantly in expectation. The results are presented in the left panel of Fig.~\ref{figDM}. As shown, the test yields a statistically significant result in only one case, indicating that, in general, the differences in predictive performance between the models are not statistically significant.
The lack of statistically significant results, despite visually noticeable differences in prediction errors, can be attributed to the properties of the DM test. Specifically, the test compares the mean difference in forecast errors relative to their variance, and the greater the variance, the more difficult it is to detect significance. In our case, the error variance is relatively high, partly due to the presence of outliers (see Table~\ref{tab2}).

{To complement this analysis, we also applied the Wilcoxon signed-rank test for paired observations, in which the test statistic is based on the ranks of paired differences and does not rely on the variance of error differentials. The results, presented in the right panel of Fig.~\ref{figDM}, show that LGBM-based models outperform HAR, BMA, and LASSO, though not RF. Moreover, LGBM32 yields more accurate forecasts than the single LGBM model.}

\begin{figure}[t]
\centering
\includegraphics[width=0.505\textwidth]{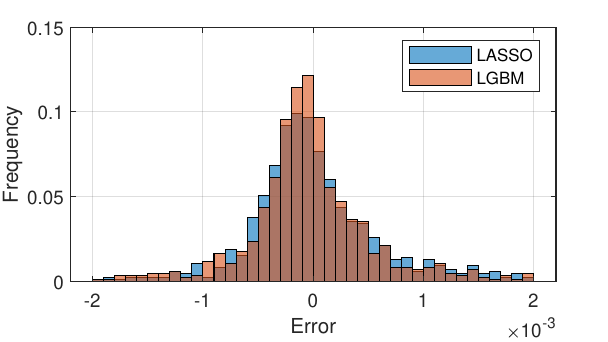}
\includegraphics[width=0.473\textwidth]{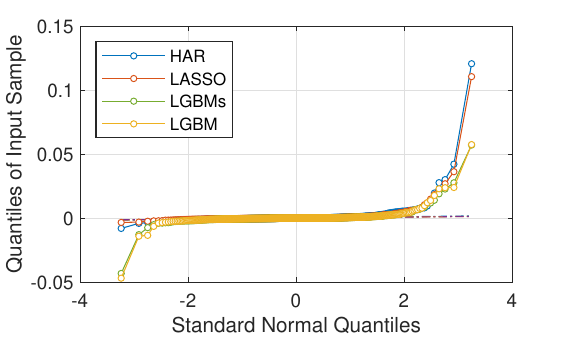}
\caption{Comparison of error distribution for selected models: histogram and quantile-quantile plot.}
\label{fig4}
\end{figure}

Fig. \ref{fig4} shows histograms of forecast errors $E$ for the LGBM model and the most accurate baseline model, LASSO. It can be observed that LGBM produces a higher concentration of forecasts with smaller errors, clustered around zero. While the distributions appear approximately symmetric, further analysis reveals that this is not the case -- extreme observations (not captured in the histogram) significantly distort the error distribution.

\begin{figure}[t]
\centering
\includegraphics[width=0.48\textwidth]{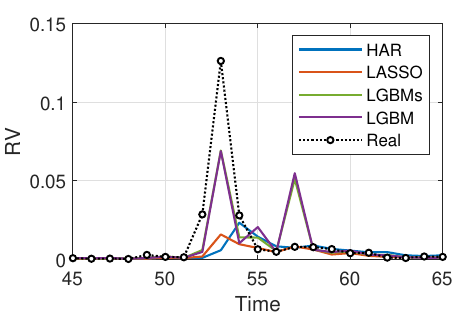}
\includegraphics[width=0.48\textwidth]{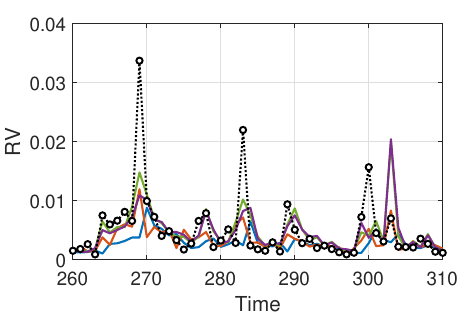}
\includegraphics[width=0.48\textwidth]{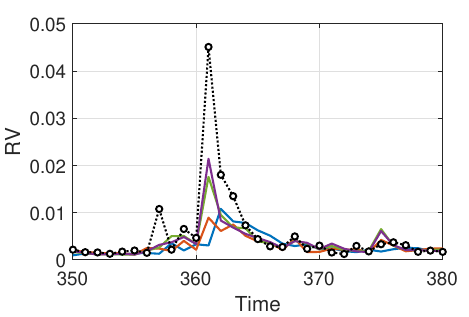}
\includegraphics[width=0.48\textwidth]{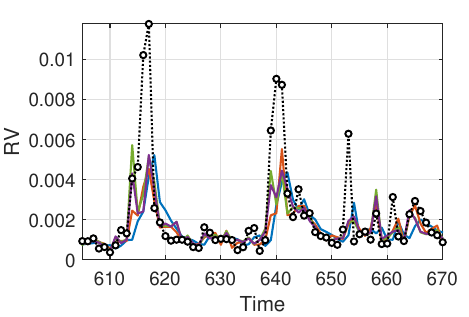}
\caption{Examples of Bitcoin RV point forecasts.}
\label{fig3}
\end{figure}

Table~\ref{tab2} presents descriptive statistics of the forecast errors $E$, where the mean error (ME) reflects forecast bias. Positive ME values indicate underprediction, which is primarily due to the nature of the RV time series that contains numerous spikes; models tend to underestimate in such scenarios. Note that LGBM-based models demonstrate more than twice the bias reduction compared to the baselines. This improvement is mainly due to better prediction of spikes (see, e.g., time points 53, 269, 283, 361, and 617 in Fig. \ref{fig3}), which is a desirable outcome. 
However, it should also be noted that LGBM-based models occasionally produce false spikes (e.g., at time points 57 and 303 in Fig. \ref{fig3}).

\begin{table}[]
\caption{Summary statistics of deterministic forecast error distribution.}
\label{tab2}
\setlength{\tabcolsep}{1pt}
\centering
\begin{tabular}{lccccccc}   
\toprule
Model   & ME& Skewness & Kurtosis & \multicolumn{4}{c}{Percentage of outliers}\\
    &&&& $< LB_{1.5}$ & $>UP_{1.5}$ & $< LB_{3}$ & $>UP_{3}$ \\
\midrule 
HAR& 5.50E-04$\pm$4.84E-03 & 19.3& 456 & 2.24    & 9.67   & 0.59  & 6.13 \\
BMA& 4.19E-04$\pm$4.81E-03 & 19.6& 464 & 2.71    & 9.20   & 0.94  & 6.01 \\
LASSO   & 4.83E-04$\pm$4.39E-03 & 19.8& 473 & 1.89    & 9.79   & 0.59  & 6.01 \\
RF & 4.84E-04$\pm$4.79E-03 & 19.4& 459 & 3.07    & 9.79   & 0.59  & 6.72 \\
LGBMs   & 1.76E-04$\pm$3.16E-03 & 5.5 & 180 & 5.31    & 8.96   & 2.95  & 5.19 \\
LGBMs32 & 1.93E-04$\pm$3.05E-03 & 10.9& 235 & 5.42    & 8.49   & 2.83  & 4.95 \\
LGBM    & 1.91E-04$\pm$3.30E-03 & 4.2 & 169 & 5.07    & 9.08   & 2.12  & 5.42 \\
LGBM32  & \textbf{1.68E-04$\pm$2.91E-03} & \textbf{1.2}& \textbf{123} & 4.83    & 9.20   & 2.00  & 5.07 \\

\bottomrule
   
\end{tabular}
  {\raggedright where $< LB_{k}$ indicates the percentage of values below the lower bound, defined as $LB_{k}=Q1-k\cdot IQR$, and $> UB_{k}$ indicates the percentage of values above the upper bound, defined as $UB_{k}=Q3+k\cdot IQR$. 
  \par}
\end{table}

High skewness values suggest systematic underprediction, driven by large positive outliers. Similarly, high kurtosis values indicate that forecast errors have heavy tails and a high frequency of extreme values, much greater than would be expected under a normal distribution. Importantly, LGBM-based models exhibit considerably lower skewness and kurtosis compared to the baseline models. Specifically, skewness and kurtosis values for LGBM32 are 1.2 and 123, respectively, while for baselines they exceed 19 and 450.

The percentage and distribution of outliers are reported in the last four columns of Table~\ref{tab2}. LGBM models produce up to 14\% outliers, compared to approximately 12\% for the baseline models. The proportion of extreme outliers reaches 8\% for LGBM and 7\% for baselines. Crucially, the distribution of outliers in LGBM models is far less asymmetric than in the baseline models: in the latter, the number of extreme negative outliers is up to 11 times smaller than the number of positive ones, whereas for LGBM models this ratio is approximately 2.5.

The Q-Q plot in the right panel of Fig.~\ref{fig4} provides a complementary view of the distribution of extreme forecast errors. It highlights a marked imbalance in the baseline models, where extreme positive errors are substantially larger in magnitude than negative ones. In contrast, the LGBM models exhibit a more symmetric distribution of extreme values, with no significant skewness toward either tail.
Note that the error distributions for LGBM models with and without shock indicators are highly similar.

\begin{figure}[t]
\centering
\includegraphics[width=0.32\textwidth]{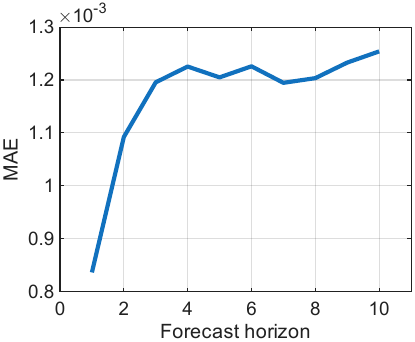}
\includegraphics[width=0.32\textwidth]{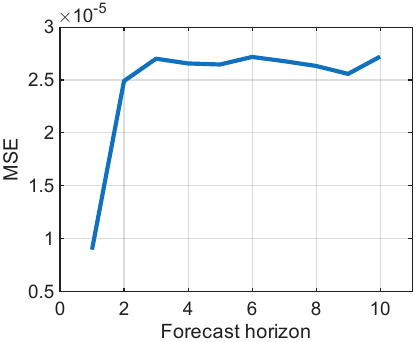}
\includegraphics[width=0.32\textwidth]{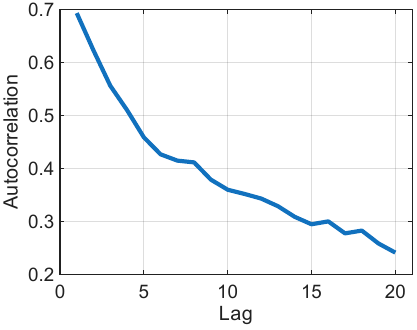}
\caption{{MAE and MSE across forecast horizons and autocorrelation of ln\_RVBTC.}}
\label{figH110}
\end{figure}

{Fig.~\ref{figH110} evaluates the forecasting accuracy of the LGBM model across horizons from $h=1$ to $h=10$. As shown, forecast errors increase sharply at $h=2$, with MAE rising by 30\% and MSE by 177\%. From horizon $h=3$ onward, the errors stabilize at approximately MAE $\approx 1.2 \times 10^{-3}$ and MSE $\approx 2.7 \times 10^{-5}$. This indicates that while the predictive power of LGBM deteriorates rapidly beyond the one-step-ahead forecast, it remains relatively stable for medium-term horizons. This pattern reflects the nature of volatility dynamics: strong short-term dependence (volatility clustering) is effectively captured at $h=1$ \cite{And03}. Supporting evidence comes from the autocorrelation plot of ln\_RVBTC in Fig.~\ref{figH110}, which shows high autocorrelation (about 0.7) at lag 1, followed by a steep decline for larger lags. This rapid decay explains the sharp increase in forecast errors at $h=2$.}

\subsection{Results for Probabilistic Forecasting}

\begin{figure}[t]
\centering
\includegraphics[width=0.47\textwidth]{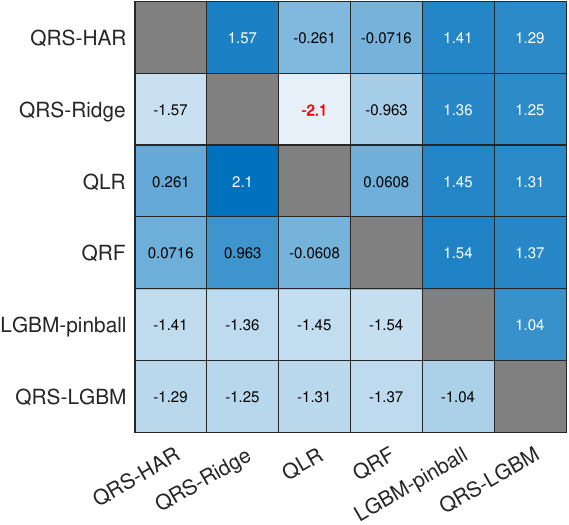}
\includegraphics[width=0.47\textwidth]{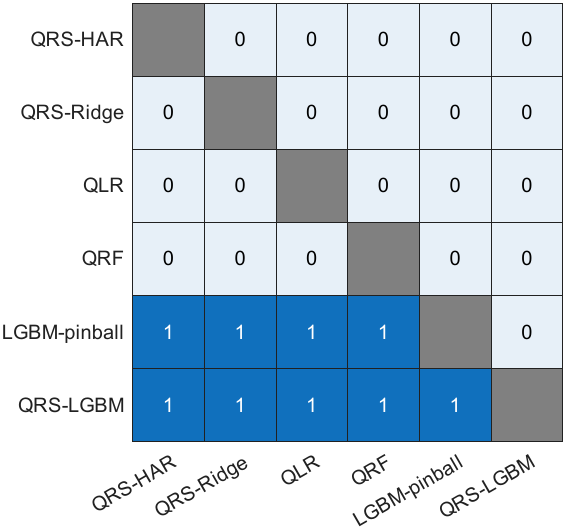}
\caption{{Diebold–Mariano (left) and Wilcoxon signed-rank (right) test results for probabilistic forecasts. In the DM test, values below $-1.96$ indicate significantly lower CRPS for the y-axis model; in the Wilcoxon test, a value of 1 indicates significantly lower errors, both at $\alpha = 0.05$.}}
\label{figDM2}
\end{figure}

Table~\ref{tab3} presents a comparison of performance metrics between the baseline models and the LGBM-based models. The best results were achieved by QRS-LGBM, followed closely by LGBM-pinball. The baseline models generally exhibited higher error levels, with one exception: QLR, which achieved a MARFE comparable to that of LGBM-pinball.
Unfortunately, as in the deterministic case, the Diebold-Mariano test did not confirm statistically significant differences in CRPS, except in one case (see left panel of Fig.~\ref{figDM2}).
{By contrast, the Wilcoxon signed-rank test confirmed the superior accuracy of LGBM-pinball and QRS-LGBM compared to the baseline models, with QRS-LGBM also outperforming LGBM-pinball (see right panel of Fig.~\ref{figDM2}).}

\begin{table}[]
\caption{Performance metrics for probabilistic forecasts.}
\label{tab3}
\setlength{\tabcolsep}{5.5pt}
\centering
\begin{tabular}{lccccc}
\toprule
Model & CRPS& MARFE    & MWS & MAE-Q    & MSE-Q    \\
\midrule 

QRS-HAR & 1.07E-03 & 7.79E-02 & 1.29E-02 & 1.31E-03 & 1.67E-05 \\
QRS-Ridge    & 1.06E-03 & 7.33E-02 & 1.25E-02 & 1.30E-03 & 1.64E-05 \\
QLR& 1.08E-03 & 4.25E-02 & 1.27E-02 & 1.34E-03 & 1.60E-05 \\
QRF& 1.10E-03 & 7.62E-02 & 1.27E-02 & 1.38E-03 & 1.67E-05 \\
LGBM-pinball & 8.86E-04 & 4.26E-02 & 8.66E-03 & 1.16E-03 & 1.04E-05 \\
QRS-LGBM& \textbf{8.13E-04} & \textbf{2.49E-02} & \textbf{6.78E-03} & \textbf{1.09E-03} & \textbf{8.06E-06} \\

\bottomrule   
\end{tabular}
\end{table}

\begin{figure}[t]
\centering
\includegraphics[width=0.49\textwidth]{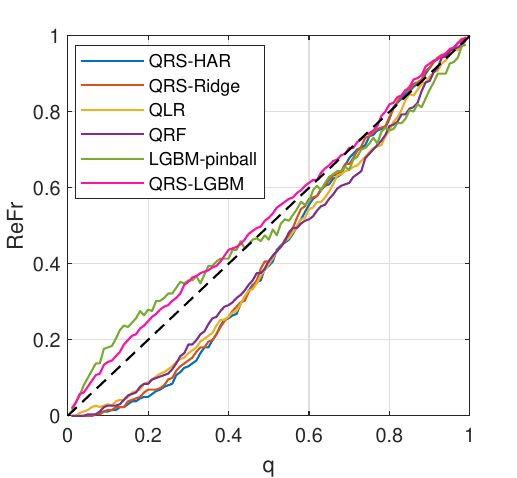}
\caption{Calibration plots.}
\label{figRF}
\end{figure}

Fig.~\ref{figRF} presents the calibration curves, with the ideal ReFr values indicated by the dashed diagonal line. As shown, the baseline models tend to underestimate quantiles, particularly for $q < 0.6$, as evidenced by the consistent downward deviation of their curves from the diagonal.
In contrast, LGBM-pinball slightly overestimates lower quantiles and underestimates upper quantiles, indicating asymmetrical calibration. The best performance is achieved by QRS-LGBM, which exhibits a slight overestimation of lower quantiles but closely approximates the upper quantiles, resulting in overall well-calibrated forecasts.
Notably, the MARFE for QRS-LGBM in Table~\ref{tab3} is 1.7 to over 3 times lower than that of the other models.

\begin{figure}[t]
\centering
\includegraphics[width=0.32\textwidth]{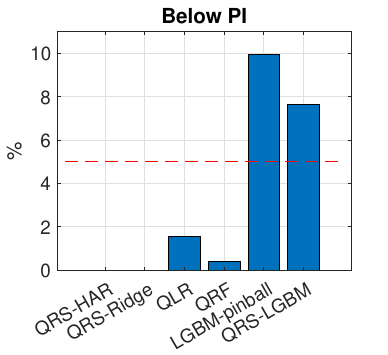}
\includegraphics[width=0.32\textwidth]{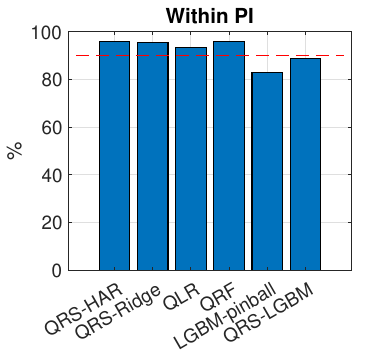}
\includegraphics[width=0.32\textwidth]{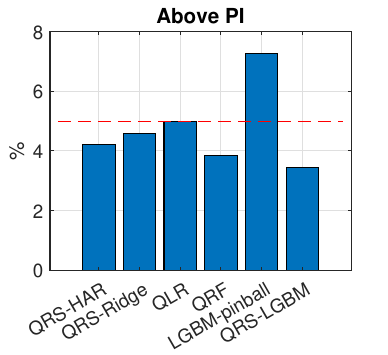}
\caption{Percentage of forecasts below, within and above the 90\% PI. The red dashed line represents the target value.}
\label{figPI}
\end{figure}

Fig.~\ref{figPI} provides a visual assessment of the 90\% PIs generated by the models.
As observed, the PIs produced by the baseline models are overly wide, with particularly low lower bounds. The MWS values reported in Table~\ref{tab3} for these models are nearly identical, confirming their similar behavior.
In contrast, the PI generated by LGBM-pinball is too narrow. The most accurate 90\% coverage is achieved by QRS-LGBM, although its PI is slightly shifted upward. Notably, the MWS for QRS-LGBM is nearly twice as low as that of the baseline models, indicating more efficient uncertainty quantification.

\begin{figure}[t]
\centering
\includegraphics[width=1\textwidth]{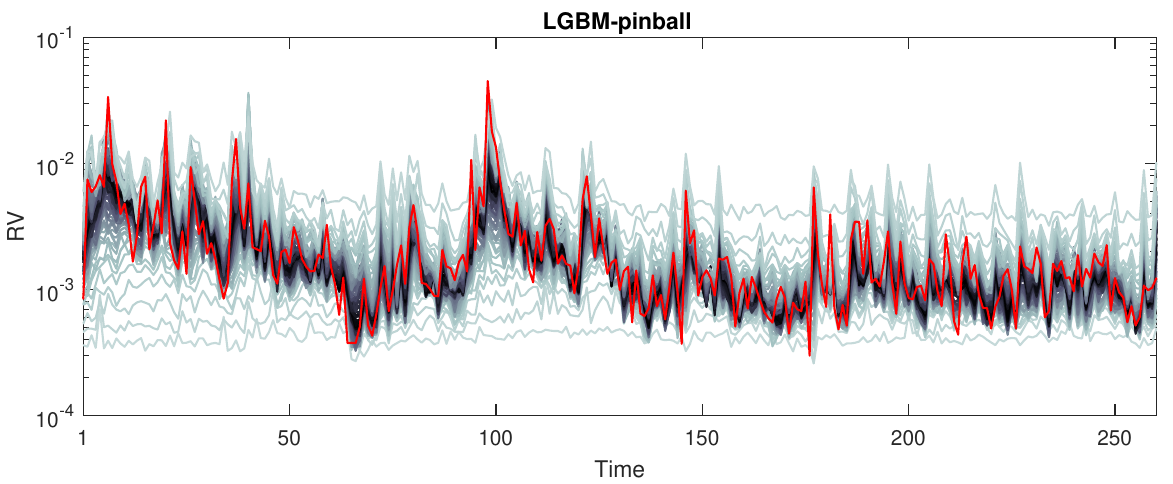}
\includegraphics[width=1\textwidth]{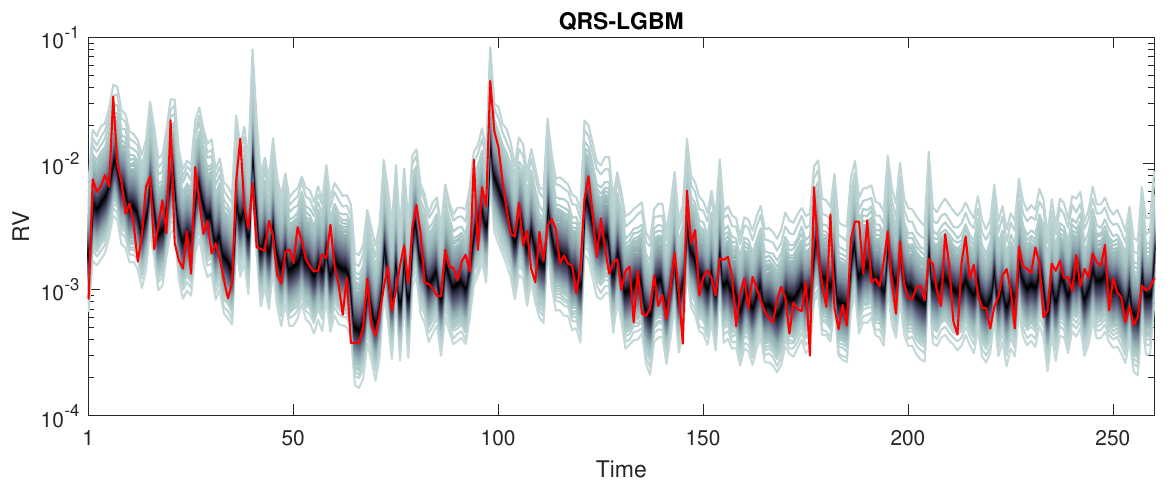}
\caption{Bitcoin RV quantile forecasts (red line represents true values).}
\label{figFor}
\end{figure}

Fig.~\ref{figFor} presents the probabilistic forecasts generated by the LGBM-based models. In the figure, quantile forecasts are represented by 99 gray lines, while the true observed values are shown in red.
One immediately noticeable pattern is that the lower quantiles predicted by LGBM-pinball tend to be lower in magnitude and exhibit less temporal variability compared to those produced by QRS-LGBM.

\begin{figure}[t]
\centering
\includegraphics[width=0.49\textwidth]{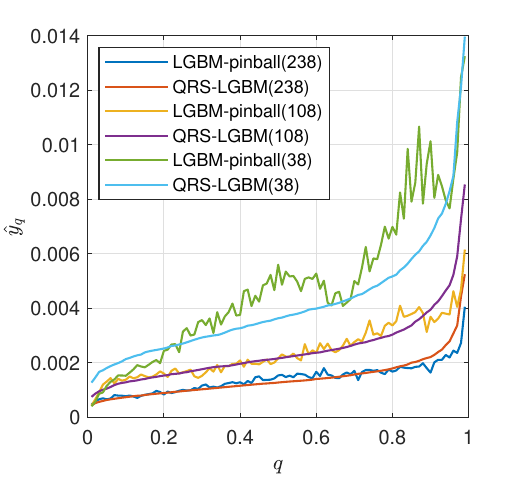}
\caption{Examples of predicted quantiles for selected test samples (sample indices shown in parentheses).}
\label{figQD}
\end{figure}

ML models trained for quantile regression may suffer from the quantile crossing problem -- a phenomenon where forecasts for lower quantiles exceed those for higher quantiles. This typically occurs when they are trained independently for each quantile level, without enforcing monotonicity constraints across quantiles.
In the case of LGBM-pinball, this issue is clearly evident in Fig.~\ref{figQD}, where the quantile trajectories appear highly irregular.
Quantile crossing was observed in over 40\% of cases for LGBM-pinball.
In contrast, the QRS-LGBM approach is inherently free from this issue, as its method of quantile construction guarantees non-crossing by design.

\subsection{Predictor Importance}

Using the two feature importance estimation methods described in Section~\ref{PrIm}, GFI and PFI, we assess the relative importance of predictors, with the goal of identifying the key drivers of Bitcoin volatility. Although both methods are derived from the structure of trained and optimized decision trees, they provide different perspectives on feature relevance.

This analysis was conducted for two deterministic forecasting models: one without shock indicators (LGBM) and one with shock indicators (LGBMs). For each model, results are reported for 848 independently trained models corresponding to each test sample.

\begin{figure}[t]
\centering
\includegraphics[width=0.49\textwidth]{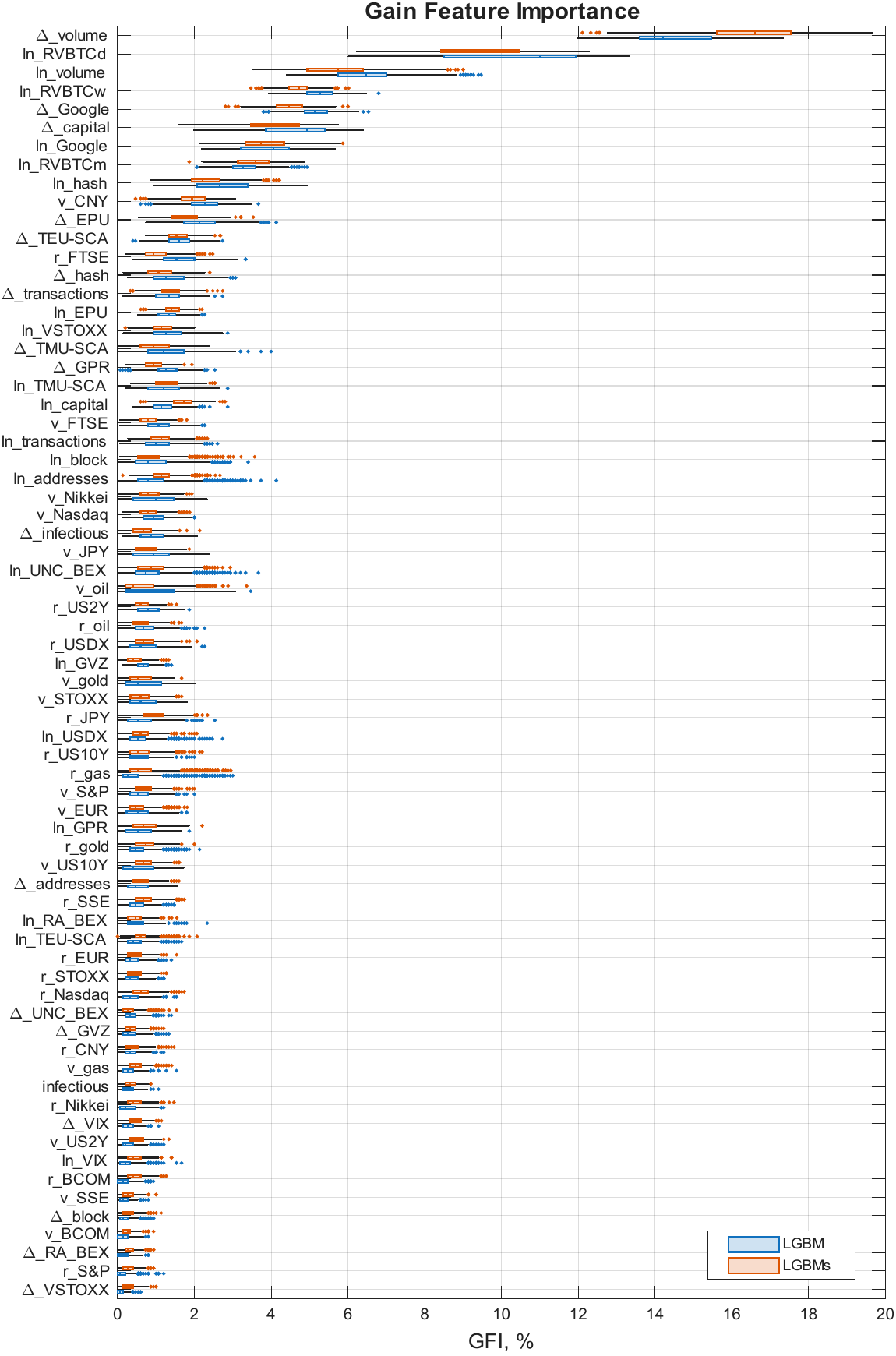}
\includegraphics[width=0.49\textwidth]{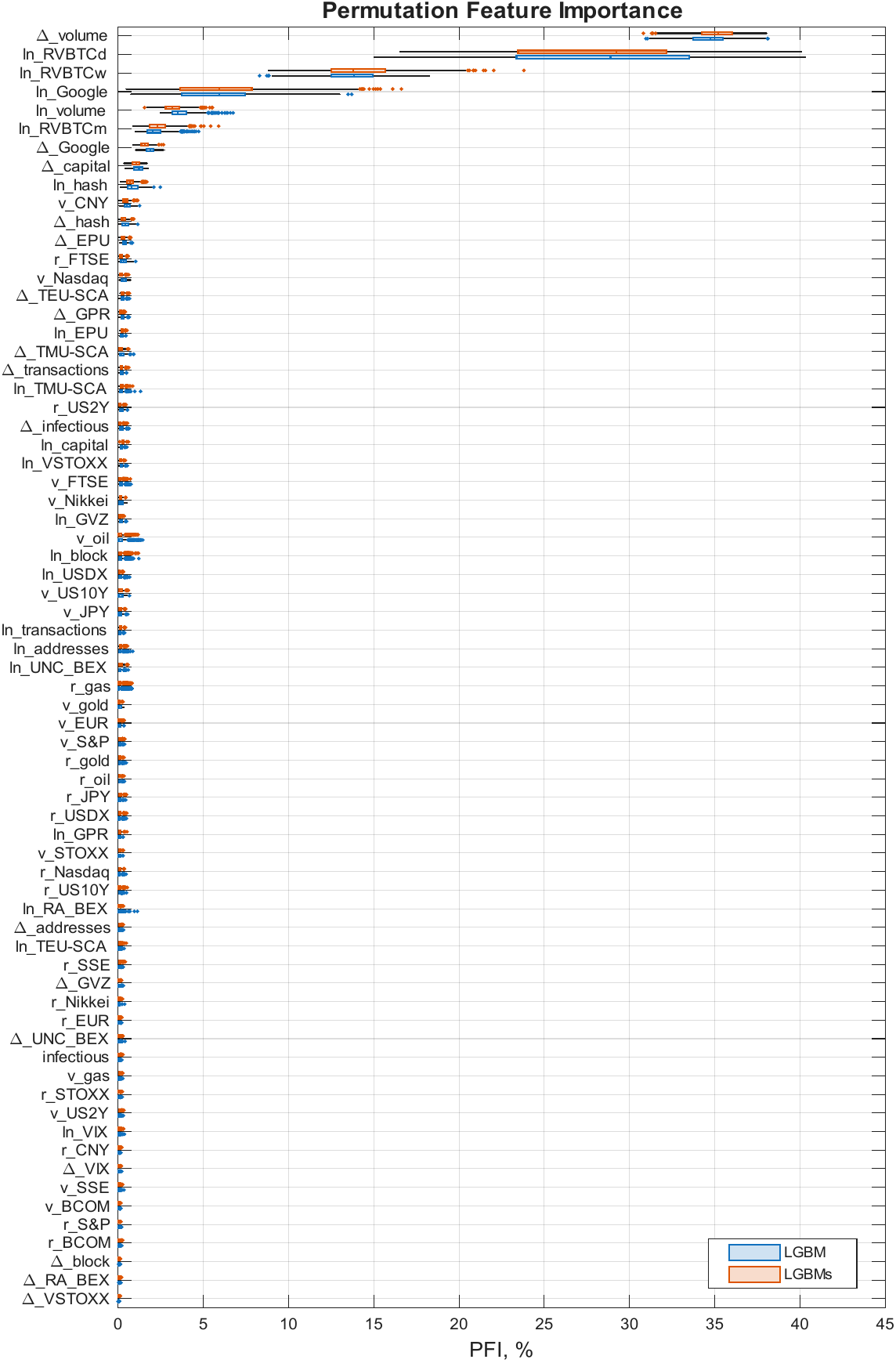}
\caption{Boxplots of GFI and PFI across 848 training sessions (without shock indicators).}
\label{figBFI}
\end{figure}

The resulting percentage values of $GFI$ and $PFI$ are shown as boxplots in Fig.~\ref{figBFI}. As illustrated, $PFI$ yields a more sharply differentiated ranking of features compared to \textit{GFI}. Specifically, $PFI$ drops rapidly from an initial value of approximately 35\%, reaching average values below 1\% by the ninth most important feature. In contrast, \textit{GFI} starts at around 15\% and maintains average values above 1\% for up to 25 predictors in the LGBM model and 18 in the LGBMs model.

The most influential predictors identified by both methods are summarized in Table~\ref{tabFI}. The key drivers of Bitcoin volatility include variables representing trading volume, past values of RV, Google search interest in Bitcoin, and market capitalization. Somewhat less influential predictors include the average daily hash rate, the CNY/USD exchange rate, and the Economic Policy Uncertainty (EPU) index derived from newspaper coverage.

\begin{table}[]
\caption{Top predictors (above 1\% importance).}
\label{tabFI}
\setlength{\tabcolsep}{2pt}
\centering
\begin{tabular}{lcccc}
\toprule
Predictor   & $GFI$ (LGBM) & $GFI$ (LGBMs) & $PFI$ (LGBM) & $PFI$ (LGBMs) \\
\midrule
$\Updelta$\_volume   & 14.48 & 16.52  & 34.61 & 35.10  \\
ln\_RVBTCd  & 10.40 & 9.53   & 28.49 & 28.36  \\
ln\_volume  & 6.50  & 5.81   & 3.66  & 3.22   \\
ln\_RVBTCw  & 5.29  & 4.70   & 13.70 & 14.21  \\
$\Updelta$\_Google   & 5.15  & 4.46   & 1.86  & 1.57   \\
$\Updelta$\_capital  & 4.61  & 4.04   & 1.17  & 1.05   \\
ln\_Google  & 3.89  & 3.81   & 5.71  & 5.95   \\
ln\_RVBTCm  & 3.32  & 3.52   & 2.23  & 2.36   \\
ln\_hash    & 2.76  & 2.32   &  &   \\
v\_CNY & 2.22  & 1.93   &  &   \\
$\Updelta$\_EPU & 2.14  & 1.75   &  &   \\
$\Updelta$\_TEU-SCA  & 1.60  & 1.57   &  &   \\
r\_FTSE& 1.60  & 1.02   &  &   \\
$\Updelta$\_hash& 1.37  & 1.09   &  &   \\
$\Updelta$\_transactions  & 1.31  & 1.38   &  &   \\
ln\_EPU& 1.30  & 1.43   &  &   \\
ln\_VSTOXX  & 1.30  & 1.13   &  &   \\
$\Updelta$\_TMU-SCA  & 1.27  &   &  &   \\
$\Updelta$\_GPR & 1.27  &   &  &   \\
ln\_TMU-SCA & 1.22  & 1.29   &  &   \\
ln\_capital & 1.20  & 1.68   &  &   \\
v\_FTSE& 1.07  &   &  &   \\
ln\_transactions & 1.04  & 1.13   &  &   \\
ln\_block   & 1.00  &   &  &   \\
ln\_addresses    &  & 1.16   &  &  \\
\bottomrule
\end{tabular}
\end{table}

The average \textit{GFI} values of the shock indicators in the LGBMs model were below 0.18\%, with approximately 94\% falling below 0.05\%. Similarly, the average \textit{PFI} values remained under 0.06\%, with about 94\% of them below 0.01\%. These results indicate that the contribution of shock indicators to the model's predictive performance is negligible. The shock variables that exhibited the greatest temporal variability were those associated with ln\_RVBTCd, $\Updelta$\_volume, $\Updelta$\_Google, and ln\_TMU-SCA (Twitter-based market uncertainty
index).

\begin{figure}[t]
\centering
\includegraphics[width=1\textwidth]{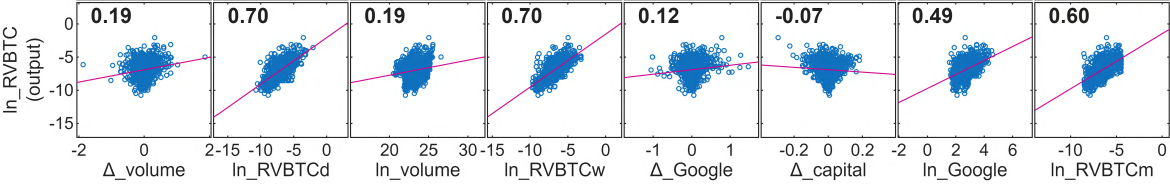}
\caption{Scatter plots of the target versus key predictors, with Pearson correlation coefficients.}
\label{figCorrFI}
\end{figure}

Fig.~\ref{figCorrFI} presents the relationships between the target variable, ln\_RVBTC, and the most important predictors. The strongest linear correlations, around 0.6-0.7, are observed between the target and its lagged/aggregated values (ln\_RVBTCd, ln\_RVBTCw, and ln\_RVBTCm). A moderate correlation of approximately 0.49 is found between the target and ln\_Google, while the change-based variable $\Updelta$\_Google shows a much weaker linear association ($\rho = 0.12$). Volume-based predictors, despite having the highest $GFI$ and $PFI$ scores, are only weakly correlated with the target ($\rho = 0.19$), and $\Updelta$\_capital exhibits virtually no linear relationship. 
However, it is important to note that low linear correlation does not preclude nonlinear dependence. Tree-based models such as LGBM are capable of capturing complex, nonlinear interactions, which may explain the high importance of these predictors despite their weak Pearson correlations.

\begin{figure}[t]
\centering
\includegraphics[width=1\textwidth]{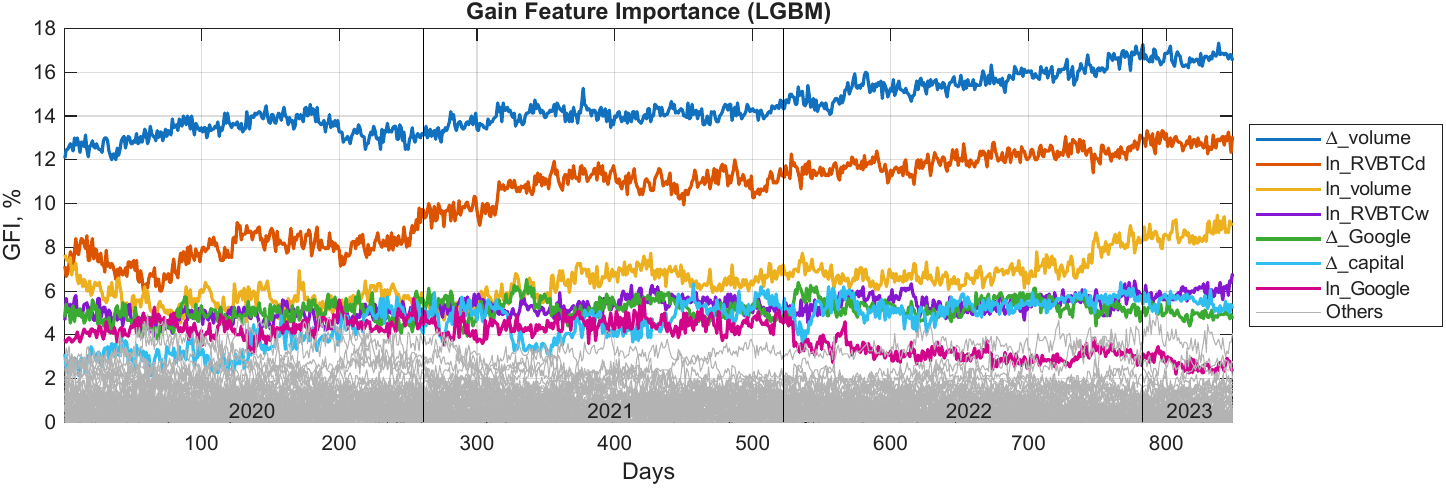}
\includegraphics[width=1\textwidth]{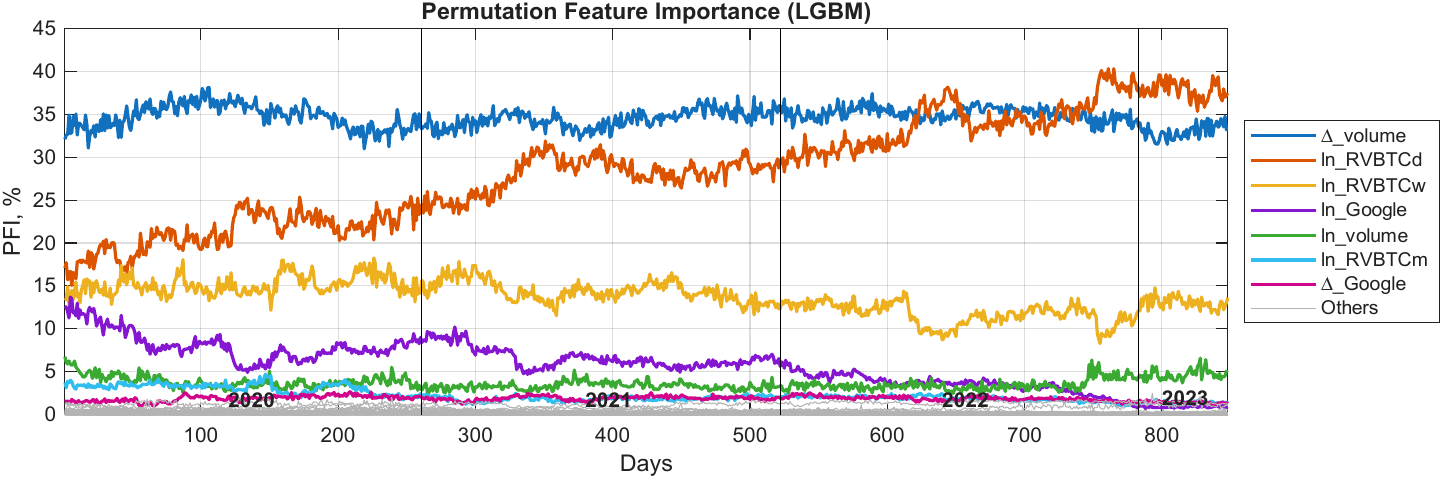}
\caption{Temporal dynamics of gain and permutation feature importance (without shock indicators).}
\label{figTFI}
\end{figure}

The temporal dynamics of \textit{GFI} and \textit{PFI}, presented in Fig.~\ref{figTFI}, illustrate how the importance of individual predictors evolved over the test period. According to \textit{GFI}, the three most influential predictors gradually increased their contribution over time, with $\Updelta$\_volume clearly dominating. In contrast, \textit{PFI} shows no upward trend for $\Updelta$\_volume; its importance remains relatively stable at around 35\%. While $\Updelta$\_volume is the dominant predictor in the early part of the period (2020-2021), it is overtaken in 2023 by ln\_RVBTCd, whose importance steadily increases throughout the test period.

Furthermore, based on \textit{PFI}, the importance of both variables related to Google search activity exhibits a declining trend. Conversely, \textit{GFI} indicates a slight upward trend for $\Updelta$\_Google, which reaches a notable 6\% by 2023.

\subsection{Importance of LGBM Hyperparameters}




After completing hyperparameter optimization using the TPE, Optuna estimates the relative importance of each hyperparameter using a post-hoc analysis based on functional ANOVA (fANOVA). This method analyzes how much each hyperparameter contributes to the variation in the model’s performance.

Key steps of this algorithm are as follows \citep{Hut14}:

\begin{enumerate}
  \item \textbf{Trial Log}: Optuna records the history of all trials, including sampled hyperparameter values and the resulting objective scores (MAE in our case).
  \item \textbf{Surrogate Model}: A random forest regressor is trained on this trial data to approximate the relationship between hyperparameters and the objective function.
  \item \textbf{Variance Decomposition (fANOVA)}: The algorithm computes the marginal contribution of each hyperparameter and their interactions to the total variance in performance.
  \item \textbf{Importance Scores}: The importance of each hyperparameter is quantified as the proportion of explained variance it contributes, normalized to percentage values.
\end{enumerate}

Higher importance means the performance of the LGBM model is more sensitive to changes in that hyperparameter. It reflects which parameters were most influential in the observed optimization space, not general theoretical importance.

Table~\ref{tabHI} presents the estimated importance of hyperparameters. As shown, \texttt{n\_estimators} is by far the most influential parameter, accounting for 83\% to 94\% of the total importance, depending on the model. For deterministic forecasting models, the second most important hyperparameter is \texttt{max\_depth}, contributing 8-12\% to the overall variance in performance. In contrast, for the probabilistic LGBM-pinball model, the importance of \texttt{max\_depth} drops significantly to 1.31\%. Other hyperparameters with importance above 1\% include \texttt{max\_bin} for LGBM and both \texttt{learning\_rate} and \texttt{bagging\_freq} for LGBM-pinball. The remaining parameters exhibit negligible importance.

\begin{table}[]
\caption{LGBM hyperparameter importance (\%).}
\label{tabHI}
\setlength{\tabcolsep}{8pt}
\centering
\begin{tabular}{lccc}
\toprule
Hyperparameter      & LGBM  & LGBMs & LGBM-pinball \\
\midrule
\texttt{n\_estimators}       & 83.55 & 90.29 & 94.27        \\
\texttt{max\_depth}          & 12.34 & 8.07  & 1.31         \\
\texttt{max\_bin}            & 1.70  & 0.20  & 0.39         \\
\texttt{min\_data\_in\_leaf} & 0.84  & 0.21  & 0.11         \\
\texttt{bagging\_fraction}   & 0.79  & 0.03  & 0.17         \\
\texttt{learning\_rate}      & 0.35  & 0.21  & 1.66         \\
\texttt{reg\_alpha}          & 0.21  & 0.13  & 0.23         \\
\texttt{feature\_fraction}   & 0.09  & 0.28  & 0.74         \\
\texttt{reg\_lambda}         & 0.07  & 0.21  & 0.06         \\
\texttt{bagging\_freq}       & 0.06  & 0.34  & 1.05         \\
\texttt{shock\_threshold}    & -     & 0.03  & -  \\
\bottomrule
\end{tabular}
\end{table}

\subsection{Discussion and Directions for Future Research}

\subsubsection{Suitability of LGBM Models for Forecasting Bitcoin Volatility}

Forecasting Bitcoin volatility is inherently challenging due to the high noise-to-signal ratio, the influence of speculative trading, extreme price swings, and its limited macroeconomic grounding. As emphasized by \citep{charles2019volatility} and \citep{Koc24}, traditional volatility models such as GARCH often struggle to accommodate the heavy-tailed and nonstationary nature of crypto assets. Moreover, the presence of volatility jumps and structural breaks, often driven by exogenous shocks like regulatory news or hacking incidents \citep{Lyo20}, calls for models that can flexibly adapt to abrupt changes and nonlinear dependencies.

The LGBM framework, due to its ability to model complex nonlinear interactions and automatic handling of missing data, offers significant advantages for forecasting Bitcoin volatility. Unlike many deep learning approaches, LGBM does not require extensive preprocessing, such as scaling or imputation 
and can natively process categorical features without one-hot encoding. It is computationally efficient and naturally incorporates ensemble learning through boosting. Nevertheless, LGBM's inability to directly handle temporal dependencies is a limitation. In our case, this was partly addressed by engineering lag-based RVBTC features. 

{\subsubsection{Overfitting Control and Robustness Enhancement in LGBM}}

{In LGBM, we reduce the risk of overfitting and enhance model robustness using three key strategies:  
\begin{itemize}
    \item \textbf{Cross-validation with multiple random seeds:} During model optimization, we employ cross-validation with several random seeds to ensure stable performance and prevent overfitting by monitoring results on validation sets. 
    \item \textbf{Strong regularization:} The hyperparameter search explicitly tunes $L_1$ and $L_2$ penalties and constrains model complexity (e.g., \texttt{max\_depth}, \texttt{min\_data\_in\_leaf}). In addition, data subsampling strategies (e.g., \texttt{bagging\_fraction}, \texttt{feature\_fraction}) further reduce the likelihood of fitting to noise.
    \item \textbf{Ensemble averaging:} Predictions from multiple runs with different random seeds are averaged, which decreases variance and improves the generalization of the final results.  
\end{itemize}  
In addition, our strategy of retraining the model for each test sample with an updated training set ensures that the model consistently leverages the most recent information and mitigates the risk of distributional shifts between training and test data.}  

{\subsubsection{LGBM Computational Cost}}

{LGBM is highly efficient in training. A single model required approximately 0.2 seconds on an Intel Ultra 9 285k processor, and training 848 separate models took only about 3 minutes.}  

{However, training 99 models for each test case in quantile forecasting remains computationally demanding, requiring roughly 5 hours for all 848 test points. To mitigate this, we implemented QRS as an alternative, which is considerably more efficient. This method requires training only one model per step and derives quantiles from historical residuals, providing a scalable solution with minimal computational overhead compared to point forecasting.}

\subsubsection{Deterministic Forecasting Performance and Role of Shock Indicators}

Our experiments confirm that LGBM-based models consistently outperform traditional baseline approaches in deterministic forecasting of Bitcoin volatility. Although the inclusion of shock indicators did not lead to substantial improvements in average predictive accuracy, it may enhance the model's responsiveness to abrupt market disruptions, potentially improving performance during extreme events.

The ensemble variants (LGBM32 and LGBMs32) demonstrated modest gains in predictive accuracy, suggesting that ensemble averaging can help stabilize forecasts under volatile conditions. However, more advanced ensemble techniques, such as stacking with meta-learning frameworks \citep{dudek2024stacking,dudek2025probabilistic} combined with deliberate diversification of base learners, hold promise for further improving model robustness and generalization.

\subsubsection{Probabilistic Forecasting Evaluation}

The probabilistic models QRS-LGBM and LGBM-pinball demonstrated superior reliability of prediction intervals compared to baseline methods. Notably, QRS-LGBM achieved better calibration and more accurate coverage probabilities, underscoring its effectiveness in capturing forecast uncertainty. In contrast, LGBM-pinball suffered from the quantile crossing issue, a common drawback when quantiles are estimated independently. QRS-LGBM avoids this problem by design, employing a structured quantile estimation procedure that enforces monotonicity across quantile levels.

The accuracy of QRS-LGBM could potentially be enhanced by refining the sampling strategy used to estimate the residual distribution. For instance, by selecting residuals from time windows immediately preceding the forecast or from observations that exhibit the greatest similarity to the query pattern.

For future research, it would be worthwhile to explore alternative GBM frameworks explicitly designed for probabilistic forecasting. One promising candidate is NGBoost (Natural Gradient Boosting), which models full predictive distributions through a parametric probabilistic framework using natural gradients. Such approaches may enhance distributional fidelity and provide more interpretable uncertainty estimates, further improving the robustness of volatility forecasts.

{\subsubsection{Economic Interpretation of Probabilistic Metrics}}

{Volatility forecasts play a central role in a wide range of financial decisions, extending beyond risk management to include spot and derivative pricing as well as portfolio allocation \cite{And03}. Accordingly, the accuracy metrics for probabilistic forecasts employed in this study -- CRPS, MARFE, and Winkler scores -- should not be viewed solely in statistical terms, but also as measures with direct relevance for financial decision-making.}

{CRPS evaluates the accuracy of the entire predictive distribution. A lower CRPS indicates that the forecast distribution is closer to the true distribution of realized volatility. In portfolio risk management, accurate probabilistic forecasts enable more reliable estimation of Value-at-Risk (VaR) and Expected Shortfall. A reduction in CRPS, therefore, implies improved tail risk estimation, supporting more robust capital allocation and margin requirements. In option pricing, where volatility is a key input, improved CRPS suggests that the model delivers more accurate volatility distributions, thereby enhancing the pricing and hedging of volatility-dependent derivatives (e.g., Bitcoin options).} 

{MARFE quantifies calibration, i.e., how closely the predicted quantiles match empirical frequencies. For risk managers, this means that confidence intervals for volatility forecasts are trustworthy, reducing the likelihood of unexpected breaches in risk limits. In trading strategies, miscalibrated forecasts can lead to mispriced volatility trades (e.g., straddles or strangles). QRS-LGBM’s superior calibration (MARFE = 0.0249 vs.\ 0.0779 for QRS-HAR) mitigates such model-induced mispricing.}  

{The Winkler Score evaluates both the accuracy and width of PIs, penalizing intervals that are either too narrow (overconfident) or too wide (uninformative). A lower MWS indicates that the model produces PIs that are simultaneously sharp and well-calibrated. In practice, this allows risk managers to rely on tighter PIs without sacrificing coverage, thereby enabling more precise capital reserves or stop-loss thresholds. For example, QRS-LGBM’s MWS of 0.00678 (vs.\ 0.0129 for QRS-HAR) implies that its 90\% PIs are nearly twice as efficient -- narrower yet equally reliable -- which directly translates into cost savings in margin requirements or option premiums.} 

{In summary, the improvements in CRPS, MARFE, and MWS achieved by QRS-LGBM are not merely statistical; they translate into tangible financial benefits: (i) more accurate risk assessments and greater capital efficiency, (ii) improved pricing of volatility-sensitive instruments, and (iii) enhanced robustness of automated trading and hedging systems that rely on volatility forecasts. Future work could include direct application case studies, such as backtesting a VaR model or evaluating options trading strategies using the proposed forecasts, to further demonstrate their economic value.} 

\subsubsection{Feature Importance Estimation: GFI vs. PFI}

GFI and PFI offer complementary perspectives on model interpretability and the identification of Bitcoin volatility drivers. GFI, computed from the internal structure of decision trees, is computationally efficient and easy to apply. However, it is known to be biased toward features with high cardinality or those used more frequently for splits, which may overstate their true predictive contribution. In contrast, PFI quantifies a feature's importance by measuring the increase in prediction error after randomly permuting its values, thus capturing its marginal impact on model performance. This method is model-agnostic and better accounts for feature interactions, but it is computationally intensive and can be unstable in the presence of multicollinearity.

Recent literature, particularly \citep{molnar2020interpretable}, highlights the advantages of SHAP (SHapley Additive exPlanations) values as a unified, theoretically grounded approach to feature attribution. SHAP offers both global and local interpretability and ensures consistency in feature ranking. In future research, it would be valuable to complement GFI and PFI analyses with SHAP-based methods to obtain a more comprehensive understanding of feature contributions.

An important advantage of our modeling strategy -- training separate models for each test sample -- is the ability to observe the temporal dynamics of feature importance. The evolution of GFI and PFI over time (see Fig.~\ref{figTFI}) provides valuable insights into how the relevance of individual predictors shifts across different market regimes. This is particularly important in the context of highly non-stationary and sentiment-driven markets such as cryptocurrencies, where the drivers of volatility can change rapidly.

\subsubsection{Identified Drivers of Bitcoin Volatility}

The feature importance analyses using both GFI and PFI consistently identified lagged RV (daily, weekly, and monthly), trading volume, Google search trends for Bitcoin, and market capitalization as the most influential predictors of Bitcoin volatility. These findings corroborate earlier studies, such as \citep{fiszeder2025bitcoin}, \citep{Bou23}, and \citep{Her22}, highlighting the central role of internal market dynamics and investor attention in shaping volatility patterns in the cryptocurrency market.

In contrast, macro-financial variables such as exchange rates (e.g., CNY/USD), economic policy uncertainty indices, and international capital flows exhibited negligible predictive power. This supports the view that Bitcoin volatility is largely decoupled from conventional macroeconomic drivers and more strongly governed by market-specific and sentiment-related factors.

The inclusion of shock indicators, designed to capture sudden changes or structural breaks in key predictors, had only a marginal impact on model performance and feature importance rankings. Most shock features exhibited extremely low average GFI and PFI values, suggesting that their contribution to volatility prediction is limited when historical dynamics and high-frequency features are already incorporated. This implies that volatility shocks may be adequately captured by existing lagged variables or that their effects are too rare or heterogeneous to be consistently informative across samples. Nonetheless, further refinement in defining shock indicators, e.g., using event-specific tagging or regime-switching frameworks, could improve their utility in future studies.




\section{Conclusion}

{This study shows that LightGBM models are highly effective for both deterministic and probabilistic forecasting of Bitcoin volatility. Using a broad set of market, behavioral, and macroeconomic predictors, LGBM outperforms econometric and random forest baselines, with probabilistic variants achieving CRPS values up to 23\% lower than benchmarks. Feature importance analysis highlights trading volume, lagged realized variances, Google search trends, and market capitalization as dominant drivers, while shock indicators and macroeconomic variables play only a minor role. LGBM’s ability to capture nonlinear interactions without heavy preprocessing, combined with its built-in feature selection, makes it a robust and interpretable tool for volatility forecasting in cryptocurrency markets.}

\section*{CRediT authorship contribution statement}
\textbf{Grzegorz Dudek:} Conceptualization, Formal analysis, Supervision, Visualization, Writing - original draft, Writing - review end editing. \textbf{Mateusz Kasprzyk:} Methodology, Software, Investigation, Writing - original draft. \textbf{Paweł Pełka:} Investigation, Validation.

\section*{Declaration of Competing Interest}
The authors declare that they have no known competing financial interests or personal relationships that could have appeared to influence the work reported in this paper.

\section*{Acknowledgements}
The authors thank Piotr Fiszeder, Witold Orzeszko, and Radosław Pietrzyk for providing the data and results for the baseline models.

\section*{Data availability}
\url{https://doi.org/10.18150/SJHAHR}

\bibliography{references_arxiv}

\section*{Declaration of generative AI and AI-assisted technologies in the manuscript preparation process}

During the preparation of this work the authors used ChatGPT, Claude and Gemini in order to improve the language and clarity of the text. After using this tool/service, the authors reviewed and edited the content as needed and take full responsibility for the content of the published article.

\end{document}